\documentclass[reqno]{amsart}
\usepackage{amsmath}
\usepackage{amsopn}
\usepackage{amssymb}
\usepackage{amsfonts}
\usepackage{graphicx}
\usepackage{boldline}
\usepackage{graphicx,booktabs}
\usepackage[table]{xcolor}
\usepackage{subfigure}
\usepackage{fancyhdr}
\usepackage[section]{placeins}
\usepackage [autostyle, english = american]{csquotes}
\usepackage{multicol}
\usepackage{booktabs,dcolumn,caption}
\usepackage{url}
\usepackage[colorlinks,allcolors=blue]{hyperref} %% optional
\numberwithin{table}{section}
\usepackage{subfigure}
\usepackage{makecell}
\usepackage{lscape} 
\usepackage{multirow}
\usepackage{longtable}
\usepackage{rotating}
\usepackage{slashbox}
\usepackage{algorithm}
\usepackage[noend]{algpseudocode}
\usepackage{booktabs}
\usepackage{xcolor}
\usepackage{soul}

\newcommand{\hlc}[2][yellow]{{%
    \colorlet{foo}{#1}%
    \sethlcolor{foo}\hl{#2}}%
}

\algrenewcommand\algorithmicrequire{\textbf{Input:}} %for algorithm- doublekaiser
\algrenewcommand\algorithmicensure{\textbf{Output:}}

\newcolumntype{R}[1]{>{\raggedleft\let\newline\\\arraybackslash\hspace{0pt}}m{#1}}
\newcolumntype{L}[1]{>{\raggedright\let\newline\\\arraybackslash\hspace{0pt}}m{#1}}
\newcolumntype{C}[1]{>{\centering\let\newline\\\arraybackslash\hspace{0pt}}m{#1}}

\usepackage[foot]{amsaddr}

\definecolor{maroon}{cmyk}{0,0.87,0.68,0.32}

\begin{document}

\title{\textbf{Semantic Analysis for Automated Evaluation of the Potential Impact of Research Articles} }
	
\author{Neslihan Suzen}
\author{Alexander Gorban}
\author{Jeremy Levesley}
\author{Evgeny Mirkes}

\address{$^{1}$School of Mathematics and Actuarial Science, University of Leicester, Leicester LE1 7RH, UK}
\address{$^{2}$Lobachevsky University, Nizhni Novgorod, Russia}

\email{ns433@leicester.ac.uk, ns553@leicester.ac.uk (N. Suzen)}
\email{ag153@leicester.ac.uk (A.N. Gorban)}
\email{jl1@leicester.ac.uk   (J. Levesley)}
\email{em322@leicester.ac.uk (E.M. Mirkes)}

\maketitle

\begin{abstract}

Can the analysis of the semantics of words used in the text of a scientific paper predict its future impact measured by citations? This study details examples of automated text classification that achieved 80\% success rate in distinguishing between highly-cited and little-cited articles. Automated intelligent systems allow the identification of promising works that could become influential in the scientific community. 
 
The problems of quantifying the meaning of texts and representation of human language have been clear since the inception of Natural Language Processing. This paper presents a novel method for vector representation of text meaning based on information theory and show how this informational semantics is used for text classification on the basis of the Leicester Scientific Corpus. 
  
We describe the experimental framework used to evaluate the impact of scientific articles through their informational semantics. Our interest is in citation classification to discover how important semantics of texts are in predicting the citation count. We propose the semantics of texts as an important factor for citation prediction. 
 
For each article, our system extracts the abstract of paper, represents the words of the abstract as vectors in Meaning Space, automatically analyses the distribution of scientific categories (Web of Science categories) within the text of abstract, and then classifies papers according to citation counts (highly-cited, little-cited). 
  
We show that an informational approach to representing the meaning of a text has offered a way to effectively predict the scientific impact of research papers.

\vspace{5mm}


\end{abstract}

\tableofcontents

\section{Introduction}

Studies in Natural Language Processing (NLP) have led to automatic and computational analysis of meaning for a large family of texts. Advances in NLP have created opportunities for the extraction of patterns in texts that can be interpreted as `meaning'. As discussed in \cite{nesli2}, we have looked at this from a holistic point of view, using simple Bag of Words (BoW) model; text is a collection of words, the meaning of text is hidden in the context of the use, which is evaluated as a whole. In \cite{CompMeaning} a `space of meaning' for words was created by measuring the extent to which a word was indicative of a category in the Web of Science (WoS). We now return to these texts and introduce a number of novel reduced representations of texts, and explore which of these is best suited for prediction of citation numbers of those papers.

In this paper, each text is a cloud of words represented in the Meaning Space (MS) \cite{nesli2}. In the MS, words are represented by their Relative Information Gains (RIG) for categories from the WoS. The MS is a space whose dimension is the number of categories in the WoS. We construct a reduced representation of each text by using properties of the distribution of words' RIGs in each category. The information in each text is summarised using a set of parameters such as mean point of a cloud's points (vector of means). A \textit{Feature Vector of Text (FVT)} is a subset of such parameters. We introduce five such FVTs by combining the mean vector, the vector of the first principal components for each text and the two centroid vectors obtained by $k$-means clustering of words in the text, with $k=2$. 

In our context, informational semantics becomes the analysis of each of FVTs for classifying texts as highly or infrequently cited: a binary classification problem. Citation count is a widely-used measure of impact in the scientific world.  As well as many other important factors that contribute to the citation counts, such as the number of scientific collaboration or the journal impact factor\cite{aksnes}, perhaps the most important factor is the context of the paper. We investigate this question under the assumption that the informational semantics in texts is an important indicator in assessment of research quality for each of the individual WoS categories.   

We hypothesize that research trends and the impact of articles can be described using semantic relations between contexts in texts. We develop and test predictive models for citation based on informational representations of texts (the various FVTs). We make the additional claim that such a linking of text semantics to scientific success in fact provides a basic tool for decisions related to publication of articles in scientific journals.    

The data we use in this study are taken from the Leicester Scientific Corpus (LSC), with citation counts extracted from the WoS website for approximately 4 years starting in 2014 \cite{nesli,nesli3,nesli4,nesli5,nesli6}. We have performed $k$-Nearest Neighbour (kNN) and Linear Discriminant Analysis (LDA) on our data and tested five different FVTs for predicting citation. In the classification task we make a very broad separation of citations: highly-cited papers and little-cited papers. Therefore, we deal with binary classification throughout this research. For experimentation, three categories are selected from three main branches of science: Applied Mathematics, Biology, and Management. The results are discussed for these categories.  

Distinguishing between the highly-cited and little-cited papers can be done in several ways. In order to avoid the effects of highly-cited categories, we use relative thresholds for each category rather than an absolute threshold defined for the whole corpus. Two rules for defining thresholds are used and two classification problems are solved. In the first task, we make a very broad separation between papers based on the average citation in the category. The classification of papers labelled as highly-cited (H) and little-cited (L) is performed. We also differentiate between extremely highly-cited (EH) and extremely little-cited (EL) papers; these correspond to the highest and lowest quartiles of the citation counts in the category respectively.
 
For both classification problems, we have applied classifiers for the five FVTs in three different instances of the MS and compare the performance of the classification.  The best classification result is achieved when combining the vectors of means and the first principal component for the category Management, and using LDA as classifier, with sensitivity and specificity both being greater than 80\%. The semantics represented by this vector allows the identification of extremely highly-cited (EH class) papers with 83.22\% accuracy (sensitivity) and the identification of extremely little-cited papers (EL class) with 81.81\% accuracy (specificity).

In this study we found that the informational semantics of a paper contains important information about the scientific success of the paper. This fact is occasionally much clearer for some scientific categories than others. Our results indicate that development of a quantitative evaluation and predictive model of citation count is possible using the informational semantics based approach proposed here.   

The rest of the paper is organised as follows. In Section \ref{texRepModel}, we review standard and widely-used text representation models in the literature. In Section~\ref{textRep} a novel text representation method is introduced and FVTs for LSC abstracts are constructed. The fundamentals of evaluation of the potential impact of research articles with classification and evaluation methods  are discussed  in Section \ref{evaluation}. In Section \ref{desStat} the pattern of citations in the LSC corpus are analysed. Given the FVTs constructed for the LSC, the methodology for classification of texts by citation count are described, and results of classification experiments are discussed in Section \ref{CitClass}. Section \ref{concDiss} concludes the paper and suggests further possible analysis.

\section{Classical Text Representation Models} \label{texRepModel}
This section reviews commonly used models for representation of texts. The underlying goal of these models is to represent texts in a reduced way that best preserves the original `meaning' of text. Most data mining algorithms work based on features extracted from the text and different representation methods create different set of features. The first thing to be addressed is how to represent a text as an input for  a data mining algorithm.
 
One of the simplest and the most commonly used text representation models is the Bag of Words (BoW). In this model, unique words in a text are extracted and the text is represented by a set of words; syntax, semantics, orders and positions of words are ignored \cite{busi}. One of the standard ways in representing texts with words is to represent  a text as word-based vector, the \textit{Vector Space Model (VSM)}, first proposed by Salton \cite{salton71,salton75}. Each unique word in the corpus defines a dimension in the space of texts. The coefficient (weight) in each component of the vector indicates the significance of the the word in that text.

In the VSM, the most basic weighting scheme is to represent a text as a Boolean vector; the weight is 1 when the word $w$ is present in the text and 0 if it is absent. Using the word count (frequency) is another common type of word weighting. This shows how many times the word occurs in the text $d$, and is  the called \textit{Term Frequency} (TF) representation ${\rm TF}(w,d)$. The more a word appears in the text, the more it is considered to be relevant for the text. Each text in the corpus becomes a vector of TFs. In addition to TF, one may take into account the distribution of the word over the corpus. A word that appears in every text is not considered to be useful for distinguishing between texts. This is measured by \textit{Inverse Document Frequency (IDF)}. The IDF is a global measure indicating the importance of the word within the entire corpus. The IDF score for a word $w$ is defined as:

\begin{equation}
{\rm IDF}(w)=1+\log\bigg(\dfrac{{\rm Number \; of \; documents \; in \; the \; corpus}}{{\rm Number \; of \; documents \; containing \;}  w }\bigg)
\end{equation} 

The less frequent a word is the corpus, the higher the IDF is. Term Frequency-Inverse Document Frequency (TFIDF) is a common refinement of TF and IDF. It reflects the importance of a word to the text by its frequency distribution over the corpus. The TFIDF score for a word $w$ in the text $d$ is calculated as follows:

\begin{equation}
{\rm TFIDF}(w,d)={\rm TF}(w,d) \times {\rm IDF}(w).
\end{equation} 
The importance increases proportionally to TF but is offset by IDF. Words that are common in every text, such as stop words (and, the, an, it for instance) have low TFIDF even though they appear many times in texts. For each text, the TFIDF is calculated for individual words in the corpus. Then, each text is represented as a vector of TFIDFs. Vectors are usually normalised to be of length 1. The calculation of TFIDF can be done in several different ways, some of which are described in \cite{busi, robertson, wu}.

In the BoW model, every word in the text is an independent potential keyword of the text and weights for words are assigned based on the frequency of words in the text and their rarity across the corpus. Words in texts are assumed to be independent of each other. The weights of words can be binary, TF or TFIDF, but such schemes need to be tested for their effectiveness for any particular data mining task. 

As well as single words, features of vectors can be a string, phrases or any concepts characterising the text. In phrase-based models, a number of phrases are identified and phrases are treated as individual features for texts.  Phrases can be formed by using different relationships between words. Two of the most popular ways to form phrases are co-occurrence and linguistic information of words. 

With the co-occurrence, two words that occur together are identified by statistical measures and this is used to form the phrases. The lexical co-occurrence of words is also used to construct semantic spaces \cite{lund,turney}. In co-occurrence approaches, the construction process is automated and no human judgement of the meaning of words is necessary. This addresses problems with the traditional semantic space approach established by Osgood \cite{osgood1,osgood2}. The first problem was that human intervention is needed to determine a set of axes that sufficiently represent the meaning of texts. The second problem was a practical problem of gathering information from human judges to determine where words are placed along these axes. This means that human input of size the number of axes multiplied by the number of words is needed. This is a huge problem with large numbers of words.

When using the linguistic information of words, we aim to capture precise syntactic word relations and phrases of two or more words can be formed \cite{berry,lewis,salton}. Such phrases can be noun phrases or clusters of words such as adjective-noun or adverb-noun. Some alternative models including combination of statistical and syntactic phrases have been proposed in the VSM \cite{stavrianou}. Salton analysed syntactic constructions in texts and assigned importance weights to the term phrases identified in order to choose phrases to use in the index for his books\cite{salton2}. However, both approaches to forming phrases result in long vectors representing the meaning of texts.  

In \cite{stavrianou}, it is suggested that more semantic information can be gathered from phrases as they give an idea about the contexts of the texts. However, Lewis \cite{lewis2} argued that single word representation carries a better statistical quality as frequencies of words in a phrase could be misleading. Even if a word appears several times, the phrase containing the word is likely to appear only once.

One inherent limitation of traditional BoW text representation for big corpora is the sparsity of high dimensional vectors. VSM representation can result in hundreds of thousands of dimensions \cite{quan,alsumait}. We usually have a huge number of features (words) with many entries of the vector for a given text being zero. For instance, even if an abstract has 200 unique words, which is approximately the average length of an abstract, it is only a small fraction of the vocabulary from the dictionary for a large text collection as a whole. In a low dimensional space, two points (texts) can seem very close to each other but are far apart separate in high dimension. Increasing the dimensionally degrades the performance of the text processing applications because of the curse of dimensionality \cite{ikonomakis}. 

It is hard to apply NLP tasks like text categorization and clustering to sparse vectors due to computational complexity and space limitation. In such a high-dimensional space, distance metrics are usually ineffective as not all features have equal importance for the text data. An appropriate number of dimensions, in other words relevant features, are required to be chosen to gain meaningful results. Global features may result in loss of information; therefore, feature relevance is needed to define locally. To extract local relevance of features, further operations are required for construction of the vector space.

Another consequence of sparsity for BoW is that IDF gives inconsistent results for rare words within a large word set \cite{church}. For high dimensional spaces, the information from TFIDF is not enough to obtain accurate weighting of words. This leads a need to propose and implement a new method of representation of words and texts, the outcome being an improvement in the performance of text mining algorithms.

\section{An Informational Text Representation Method} \label{textRep}
This section presents an overview and a formalization of a novel text representation method for the study of extracting the `meaning of text' in a scientific corpus. We will explore the efficacy of this method on various text analysis tasks. 

Consider a text space consisting of all texts in a corpus, each containing words that are represented according to their importance extracted from their usage in subject categories. Recall that a word in the Meaning Space (MS) is represented as a vector of Relative Information Gains (RIGs) from word to category, where the dimensions of the word representation are the number subject categories. A text is a collection of words that are points in the MS. Each text is actually a cloud of word points and each cloud has a distribution of different words in the MS.

For a text, we expect that RIGs of words in any two categories (two dimensions) are not the same and not equally distributed. In a more precise sense, if the words in two texts have different degrees of importance for a particular category, the distributions of their RIGs in the category will be statistically different. We hypothesize that the central tendency and the statistical dispersion of any two clouds of such words will not be the same in the category given. In other words, for different texts the means and the variances of the distributions of RIGs in the category are statistically different. We expect that two texts with similar words are represented by points that are closely located in the MS, which means the two texts have similar `meaning'. This formally establishes a new model to represent texts' meanings. 

Now, each text has a distribution of words' RIGs in each dimension of the MS, defined by sets of parameters such as the mean vector and the vector of standard deviations. To represent a text, we introduce the \textit{Feature Vector of Text (FVT)} which stores features we expect to be significant in differentiating one text from another. The richest FVT we consider is defined as 
\begin{equation}\label{eq:FVT}
{\rm FVT}(d)=(\overrightarrow{\mu}(d) , \overrightarrow{\sigma} (d), \overrightarrow{PC_{1}}(d),...,\overrightarrow{PC_{n}}(d),\overrightarrow{c_{1}}(d),\overrightarrow{c_{2}}(d)),
\end{equation}
where $d$ is a text in the corpus, $\overrightarrow{\mu}$  is the vector of mean values of a cloud's points, $\overrightarrow{\sigma}$ is the vector of standard deviations of a cloud's points, $\overrightarrow{PC_{n}}$ is the $n^{th}$ principal component, and $\overrightarrow{c_{1}}$ and $\overrightarrow{c_{2}}$ are the centroids of the two clusters (centroid vectors) obtained by $k$-means clustering of words in the text \cite{wunsch}, with $k=2$. It is obvious that the length of each vector in the ${\rm FVT} (d)$ is equal to the number of categories.

We note that the centroids $\overrightarrow{c_{1}}$ and  $\overrightarrow{c_{2}}$ in $k$-means clustering are initialised as the two most distant points in the cloud under the assumption that two distant words should belong to different clusters (Algorithm \ref{k-means}). It is important that we do not make a random initialisation, as this would lead to different representations each time the algorithm is run. 
\begin{algorithm}
\caption{Extracting the centroids of two clusters of words for each text by $k$-means clustering}\label{k-means}
\begin{algorithmic}[1]
 \Require Text and clouds of words for each represented in the Meaning Space; the number of clusters $k=2$
 \Ensure Centroids of two clusters of words for input text
\State Initialize centroids of two clusters as two the most distant words in the cloud
\State Calculate the distance between data points and centroids
\State Form two clusters with data points assigned to nearest centroid

\Repeat
\State Re-calculate centroids based on the current partition 
\State Assign each data point to the nearest centroid   
\Until{there is no change between two iterations, that is, the algorithm is converged.}
\State \textbf{return} Two clusters and their centroid vectors.
\end{algorithmic}
\end{algorithm}

Depending on the task, the FVT can be constructed with a reduced set of features, for instance:

  \begin{equation}
  \begin{split}
 {\rm FVT}(d)&=(\overrightarrow{\mu}(d)) \\
{\rm FVT}(d)&=(\overrightarrow{\mu}(d) ,  \overrightarrow{PC_{1}}(d))  \\
{\rm FVT}(d)&=(\overrightarrow{\mu}(d) , \overrightarrow{c_{1}}(d),\overrightarrow{c_{2}}(d)).
\end{split}
\end{equation}

Since each of the vectors in the FVT is of length $n$, the number of categories in the corpus, if we use all $n$ principle components, plus the four extra categories, we use $n(n+4)$ pieces of information to represent the documents in the corpus. Of course, by choosing fewer vectors in FVT we can significantly reduce dimension. 

Given the FVT vectors for two texts, it is also possible to compute the similarity between them in a lower dimensional space, which is one of the main problems in typical text mining and Natural Language Processing (NLP) tasks, including text categorization, text clustering, concept/entity extraction, sentiment analysis, entity relation modelling and many more.

\subsection{Representation of LSC Abstracts as Feature Vector of Text (FVT)\nopunct}\hspace*{\fill} \\\\
In this subsection, we represent each text $d$ in the LSC as a $FVT(d)$ obtained from its words that are points in MS, where dimensions are 252 Web of Science categories. 

Recall that we introduced  the informational space of meaning for scientific texts in \cite{nesli2}. In this work, we prepared the Meaning Space for the LSC and each word of the LScDC is represented as a vector of RIGs calculated from the word to category. The dimension of the MS was then evaluated and PCA was applied to explore the actual number of dimensions of the MS and provide a detailed understanding of categories in the MS. 

Words can be represented in two ways in the Meaning Space: by their RIGs or in a reduced dimension using principle component analysis (PCA). In the second case, the dimension is 61 by the Kaiser rule, 16 by the Broken Stick rule or 13 by PCA-CN \cite{CompMeaning}. Therefore, texts will have distribution of words represented in 13, 16 or 61 dimensional space. 

With the words represented in any of these vector spaces, we construct the ${\rm FVT}(d)$ for each text $d$ as described below:

\begin{equation}
  \begin{split}
{\rm FVT}_{1}(d)&=(\overrightarrow{\mu}(d)) \\
{\rm FVT}_{2}(d)&=(\overrightarrow{PC_{1}}(d))  \\
{\rm FVT}_{3}(d)&=(\overrightarrow{\mu}(d) ,\overrightarrow{PC_{1}}(d))  \\
{\rm FVT}_{4}(d)&=(\overrightarrow{c_{1}}(d),\overrightarrow{c_{2}}(d))\\
{\rm FVT}_{5}(d)&=(\overrightarrow{c_{1}}(d),\overrightarrow{c_{2}}(d),\overrightarrow{PC_{1}}(d)).
\end{split}
\end{equation} 

If words are represented by RIGs vectors in the MS, the dimensions of ${\rm FVT}(d)$ are 252, 504 or 756. If the space of words is 13-Dimensional reduced basis, the dimensions of ${\rm FVT}(d)$ are 13, 26 or 39.

\section{Evaluation of the Potential Impact of Research Articles through Semantics} \label{evaluation}
The corpus of scientific texts is a dynamic resource which has changed and improved as new topics in science are introduced. A rapid evolution of research and new priorities among the world's scientists requires the continuous analysis of the corpora of research literature in order to understand the impact of research articles for understanding the development of science. Citation count is a widely used indicator for how much interest an article or topic has attracted interest in the scientific world. The context of an article is often the most important factor in deciding the citation count. It is, indeed, not easy to track the context of texts (or meaning of text) and so quickly monitor and generate trends in a growing volume of publications. 

%burdan aldim 

With the construction of the triad - dictionary, texts and representation of texts in the Meaning Space as described in the previous section, we now evaluate the scientific impact of articles via their informational semantics. Predictive models for citation based on this representation will be assessed for articles published in WoS database in 2014 with citation counts in the subsequent (approximately) 4 years. Our focus is to explore how efficiently the semantics of scientific articles can be used to predict the impact of articles as measured by citation.

In this context, we employ two classification algorithms to analyse the predictability of citation: Linear Discriminant Analysis (LDA) and $k$-nearest neighbours (kNN). In classification of impact, various criteria for distinguishing classes of papers can be used to define citation ranks. We consider this problem as a very broad separation of papers: highly-cited papers and little-cited papers, that is, binary classification of impact.  

An important element in the selection of groups of papers is the definition of citation counts for a highly-cited paper. This is controversial in research assessments. Ordinary  citation counts for certain fields can be high compared to other fields.  Utilising an absolute threshold to distinguish the classes of papers in a corpus  might lead to dominance of  papers in highly-cited scientific categories \cite{aksnes}.  To avoid this disciplinary difference in average citation within categories, one can use relative thresholds for each category. In this research, category-based relative thresholds will be defined.

\subsection{Methods of Classification\nopunct}\hspace*{\fill} \\\\
This subsection provides the methodology for classification algorithms used in the evaluation of the scientific impact of articles.  

\subsubsection{Criterion for evaluating the performance of classification methods\nopunct}\hspace*{\fill} \\\\
Data imbalance is encountered in many classification problems, we have an uneven distribution of sample size in the data classes. In particular, citation distributions are usually unequal for the classes of highly-cited and little-cited papers. The majority of papers are little-cited whereas some papers are extremely highly-cited in certain scientific disciplines, for instance medicine \cite{aksnes,aksnes2}. This leads classifiers to be inherently biased toward the class with larger size. Therefore, maximising the measure of performance such as accuracy might be misleading for imbalanced data. In this case, we need to use metrics which adequately take the class distribution into account. 

In this work, we use standard metrics to measure the performance of classifiers: sensitivity and specificity \cite{chapman}. The $2 \times 2$ confusion matrix showing correct predictions and types of incorrect predictions for binary classification is used to calculate sensitivity and specificity (Table \ref{table:contingency}). The confusion matrix presents the decision made by the classifier in four categories: True positives (TP) referring to correctly labelled as positive, False positives (FP) referring to negative samples incorrectly labelled as positive, True negatives (TN) referring to negatives correctly labelled as negative and False Negatives (FN) referring to positive examples incorrectly labelled as negative \cite{davis} . 

\begin{table}[bth]
\centering
\caption{Confusion Matrix used to calculate sensitivity and specificity for binary classification }
\renewcommand\arraystretch{1}	
\begin{tabular}	{L{2cm}| R{3cm} R{3.2cm} |R{1.5cm}}

&\multicolumn{2}{C{6.2cm}|}{\textbf{True Condition}} & \\ \hline

   &\multicolumn{1}{C{3cm}}{\textbf{Positive}} &\multicolumn{1}{C{3.2cm}|}{\textbf{Negative}} &\multicolumn{1}{C{1.5cm}}{\textbf{Total}} \\ \hline

\textbf{Positive}&	True Positive (TP)	&		False Positive (FP)	&	TP+FP  \\ 
&	 	&		&	 \\ 
\textbf{Negative	}&	False Negative (FN)	&		True Negative (TN)	&	FN+TN  \\ \hline
\textbf{Total}	&	TP+FN	&		FP+TN	&	  \\ 
\end{tabular}
\label{table:contingency}
\end{table}

Sensitivity (True Positive Rate) refers to the probability that a highly-cited paper is classified as highly-cited, calculated as:

$$ {\rm Sensitivity} = \frac{TP}{TP+FN}$$

Specificity (True Negative Rate) refers to the probability that a little-cited paper is classified as little-cited, calculated as:

$$ {\rm Specificity} = \frac{TN}{TN+FP}$$

In other words, the sensitivity is the proportion of correctly identified positives and the specificity is the proportion of  correctly identified negatives. To evaluate the performance of classifiers, we use the criterion: maximise the sum of sensitivity and specificity. The higher the sum means the better performance of the classifier. 

The ROC (receiver operating characteristic) curves are also used to present results for binary classification problems \cite{hanley}. We plot the ROC curve showing the sensitivity and the corresponding specificity points for every possible cut-off for the set in classification. The $x$-axis shows 1-specificity (false positive rate) and the $y$-axis shows sensitivity (true positive rate) for a chosen cut-off. A diagonal ROC curve means that the model predicts randomly. 

The area under the curve (AUC) is used as a criterion to measure the performance of a binary classifier: how good is the classifier in identifying the correct classes of articles. For a diagonal line, the AUC is 0.5. The higher the AUC, the better the model is at distinguishing between classes. The maximum value of AUC is 1 which indicates a perfect classifier.    

%In other words, it shows how much the classifier is capable ti  

\subsubsection{k-Nearest Neighbour (kNN)\nopunct}\hspace*{\fill} \\\\ 
K-nearest neighbours (kNN) is a simple and easy-to-implement classification method. For a test sample to be classified, the set of the $k$ nearest neighbours are found from all the training data according to a distance metric (proximity). Every data point in the training set belongs to one class or the other, and class labels of this set are known. For the given test point where the class label is unknown, kNN searches for the closest $k$ nearest neighbours, based on the distance metric, and decides the classification by the majority voting among data points in the neighbourhood \cite{cai}. We use the Euclidean distance as proximity measure. 

The value of $k$ should be defined before applying kNN and the efficacy of the classification is usually dependent on this value chosen. One of the simplest ways to determine the value of $k$ is to run the algorithm several times with different values of $k$, and to chose the optimal value of $k$ \cite{guo,arif}. The optimal value refers to the $k$ with the best classification performance. 

kNN suffers from imbalance of the data among classes as for many other classification tasks \cite{tan}. In the kNN rule, the majority category is very likely to have more samples in the set of $k$-nearest neighbours for a test sample, and so the test sample tends to be classified as the majority class. This results in high accuracy for the majority class and low accuracy for the minority class. An intrinsic method which mitigates against the effect of sample size in classes is to assign weights (inversely proportional to class frequency) to the neighbours. 

To improve the classification performance of kNN, we implement a weighting described as follows. For a given test point, we form the set of $k$-nearest neighbours from the training texts where Class 1 has $N_{1}$ samples and Class 2 has $N_{2}$ samples out of the $N$ papers. Suppose we have $k=t$ and we obtain $s$ papers from the Class 1 and $t-s$ papers from Class 2 in the decision set by kNN. If $P_{1}=N_{1}/N$ and $P_{2}=N_{2}/N$ we weight a data point in Class A and Class B with coefficients $1/P_{1}$ and $1/P_{2}$ respectively. If there are $s$ cases from Class 1 and $t-s$ cases from Class 2 in the decision set, we give Class 1 a score of $s/P_{1}$  and Class B a score of $(t-s)/P_{2}$. We assign the test point to the class that has the higher score.

\subsubsection{Linear Discriminant Analysis (LDA)\nopunct}\hspace*{\fill} \\\\
In this paper, Fisher's linear discriminant is used for the binary classification problem for distinguishing between highly-cited papers and little-cited papers \cite{fisher}. The means $ \mu_{1} $ and $ \mu_{2} $ of each class for the training set are computed. Then, covariance matrices $\Sigma_{1}$ and $\Sigma_{2}$ for each class are calculated. The direction of the discriminating line is defined as:

$$\omega = (\Sigma_{1}+\Sigma_{2})^{-1}(\mu_{1}-\mu_{2}).$$

Projections of each document $d$ on the discrimination direction are found by dot product $(\omega,d)$ \cite{awaz}. As mentioned before, the optimal threshold for the LDA is that which maximises the sum of specificity and sensitivity. This means that the decision criterion for a point being in one of the class becomes a threshold on $(\omega,d)$ that maximises the sum of sensitivity and specificity. We first apply the LDA to the whole set with two classes and calculate the sensitivity and specificity based on the threshold found.

In order to test the statistical power and the quality of the classifiers, we also test LDA with Leave-One-Out Cross Validation (LOOCV) \cite{gelman}. LOOCV is performed as follows: (1) Select a single-item as a test point; (2) Use all other points as a training set; (3) Apply the learning algorithm to training set once for each test point. The performance of the classifier with LOOCV is also evaluated using the sum of sensitivity and specificity. 

\subsubsection{Supervised PCA\nopunct}\hspace*{\fill} \\\\
For a classification problem, supervised PCA searches for a low dimensional linear manifold where the distances between projection points of different sets are maximal and the distances between projection points of the same set are minimal \cite{supPCA1,supPCA2}. Given set of points $X=(x^{1},...,x^{n}), x^{i}=(x_{1}^{i},...,x_{m}^{i}), x_{k}=(x_{k}^{1},..., x_{k}^{n}) $ where each row $x^{i}$ in the matrix $X$ corresponds to one text. Points are centred where $ \sum_{i=1}^{n}{x_{k}^{i}}=0$ for all $k=1,...,m$. $L_{1}$ and $L_{2}$ are two sets of points where those with label 1 belong to $L_{1}$ and those with label label 2 belong to $ L_{2}$.

We consider the column vectors $V=(v^{1},...,v^{p})$ with $(v^{i},v^{j})=\delta _{ij}$ where $\delta _{ij}=0$ if $i\neq j$ and $\delta _{ii}=1$ (Kronecker delta). Projection of points onto the $p$-dimensional subspace is defined as $P_{v}(x)=xV$. The squared distance between projections of two points $x^{i}$ and $x^{j}$ is

$$\parallel P_{v}(x_{i})- P_{v}(x_{j}) \parallel ^{2}=\parallel x^{i}V-x^{j}V \parallel ^{2}.$$
For two sets of points with different labels, the averaged squared distance  between projections is

$$D_{B}=\dfrac{1}{\vert L_{1} L_{2} \vert} \sum _{i \in L_{1}} {\sum_{j\in L_{2}} {\parallel P_{v}(x_{i})- P_{v}(x_{j}) \parallel ^{2}}}.$$
The averaged squared distance within a set of points with a class label is 
 
$$D_{w_{k}}=\dfrac{2}{\vert L_{k} \vert \vert (L_{k}-1) \vert} \sum _{i,j \in L_{k}, i>j} {\parallel P_{v}(x_{i})- P_{v}(x_{j}) \parallel ^{2}}.$$

Therefore, the function of interest to be maximised is 
$$ D_{C}=D_{B} - \dfrac{ \alpha }{2} (D_{w_{1}} + D_{w_{2}})$$
where $\alpha$ is a parameter that indicates the relative importance of $D_{B}$ and $D_{w_{1}} + D_{w_{2}}$.

\section{Descriptive Statistics of Citations in the LSC} \label{desStat}
In this section, descriptive statistics are used to summarise the Web of Science abstract database with respect to citation counts. They describe various aspects of the citation counts in the data, information which can be used in further studies. 

In the WoS, two types of citation are accessible: \textit{Times Cited in WoS Core Collection} and \textit{Total Times Cited}. In the first case, the count shows the total number of times a paper was cited by other papers in WoS Core Collection \cite{wosCite}. The total times cited displays the number of times the paper was cited by other items in WoS products including the WoS Core Collection. The counts of total times cited are used in this study. Recall that the dataset consists of 1,673,350 abstracts or proceeding papers that were published in Web of Science database in 2014 with citation counts in more or less the 4 years following publication (2014-2018).  

\begin{figure}[htb]
  \centering
  \includegraphics[height=.28\textheight, width=.9\textwidth]{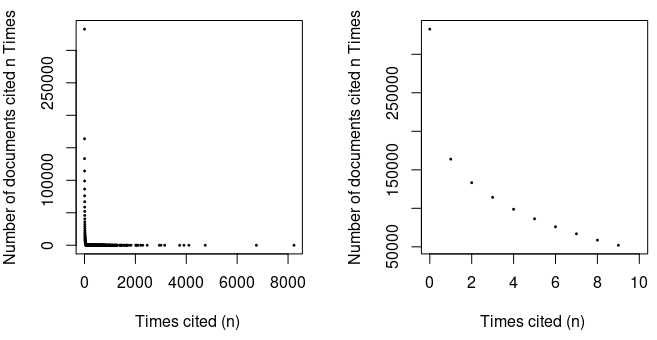}
  \caption{Graph of the number of documents cited $n$ times in the corpus LSC. The figure on the right hand side shows the number of documents cited up to 10 times. The average citation in the corpus is 9.38.}
  \label{fig:timesCited}
\end{figure}

Frequency statistics is the first descriptive statistic used to determine the distribution of citations in the corpus. Raw counts of documents for each citation count (times cited) is presented in Figure \ref{fig:timesCited}. It should be stressed that as categories are not exclusive, a citation record for a document can appear in multiple categories. However, in Figure \ref{fig:timesCited}, within a category, we only count the document once, no matter how many times it was cited. This is true of all other calculations of descriptive statistics for the corpus. Note that only 64 documents documents are cited more than 1000 times. Another 243 documents are cited from between 500 to 1000 times. However, 929,361 documents are cited no more than 5 times. In Table \ref{table:topBottom} one sees the highest 5 and the lowest 5 cited counts, supporting the fact that more than 50\% of documents cited less than 4 times. The average of citation counts is approximately 9.

\begin{table}[bth]
\centering
\caption{The highest and the lowest five citations in the corpus with the number of documents for the corresponding citations}
\renewcommand\arraystretch{1}	
\begin{tabular}	{|R{2.5cm}| R{2.5cm} |R{2.5cm}| R{2.5cm}|}
\hline
  	  \multicolumn{1}{|C{2.5cm}|}{\textbf{The highest citations}} &\multicolumn{1}{C{2.5cm}|}{\textbf{Number of Documents}} &\multicolumn{1}{C{2.5cm}|}{\textbf{The lowest citations}} &\multicolumn{1}{C{2.5cm}|}{\textbf{Number of Documents}} \\ \hline
8,234	&	1	&		0	&	332,610  \\ \hline
6,756	&	1	&		1	&	163,907  \\ \hline
4,744	&	1	&		2	&	133,280  \\ \hline
4,102	&	1	&		3	&	114,309  \\ \hline
3,908	&	1	&		4	&	98,843  \\ \hline
\end{tabular}
\label{table:topBottom}
\end{table}

It should be mentioned that citation counts may vary differently across scientific categories and so descriptive statistics of citations in individual categories may be characteristic for different branches of science. Also, the number of papers per category may be a factor that correlates with citations. This question  was also posed and analysed in \cite{patience}. In that study, the most cited 500 papers of each category (236 WoS categories) with citation counts from 2010 to 2014 were compiled. It was concluded that the citation counts are the highest in multidisciplinary sciences, general internal medicine, and biochemistry, and the lowest in literature, poetry and dance. The correlation of number of papers assigned to a category with the citation counts for the top papers selected in a category was discussed. Therefore, a detailed analysis for each category will give better insight into understanding the factors contributing to citation and help in the selection of sample representative of the corpus. 

A sample of the corpus should be selected in such as way as to be representative of the whole corpus. An example of corpus sampling is as follows. Before sampling, we consider a division of the corpus into groups of the main branches of science - social science (e.g. sociology, law and psychology),  physical science (e.g. physics and chemistry), life science (e.g. biology, medicine and physiology), earth science (e.g. geoscience, astronomy), and formal science (e.g. mathematics, computer science and logic). Then the corpus is sampled within each branch (or stratum). A sample of categories from each stratum is collected by ensuring the presence of key categories within the sample. We ensure that the sample reflects the characteristics of the corpus, assuring that the all branches will be included and appropriately represented in the sample. 
 
\begin{figure}[htb]
  \centering
  \includegraphics[height=.28\textheight, width=.9\textwidth]{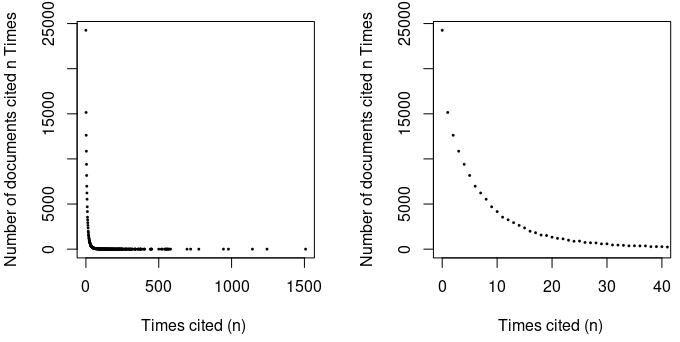}
  \caption{Graph of the number of documents cited $n$ times in the sample selected. The figure on the right hand side shows the number of documents cited up to 40 times. The average citation in the sample is approximately 9.43.}
  \label{fig:allSamplePlot}
\end{figure}

Taking into account the considerations above, we can select a sample with the categories presented in Table \ref{table:sampleSt}. One can see that the sample size of each stratum is similar, containing  approximately 30,000 documents. The total number of documents in the sample is 147,901. The counts of documents for each citation number (times cited) for the sample is given in Figure \ref{fig:allSamplePlot}. The figures shows similar trends as for the corpus with a very close value of average citation: 9.43. A more detailed explanation and statistics for each stratum and its sub-branches are provided later in this section.  

\begin{table}[h]
		\centering
		\caption{An example of corpus sampling: Sample structure}
			\begin{tabular}{ | L{1.5cm} | L{6.5cm} | R{2.5cm}|  }
	  	 	  \hline
	  	 	  \multicolumn{2}{|C{8cm}|}{\textbf{Branches and categories}} 				& \multicolumn{1}{C{2.5cm}|}{\textbf{\# of documents}} 			\\ \hline
	  	 	  
 			  											 			& Biology				& 	9,917 			\\ 
 			   \multicolumn{1}{|L{3cm}|}{Life Science}				& Medicine, Research \& Experimental			&	19,744			 	\\ \hline

 			  										 				& Psychology  &  6,989				\\ 
 			   \multicolumn{1}{|L{3cm}|}{Social Science}				& Political Science &	5,106			\\
 			       												    & Management &	14,339		\\ 
 			       													& Sociology &	4,725								\\ \hline

 			       						                             & Physics, Particles \& Fields  &	13,203			\\ 
 			   \multicolumn{1}{|L{3cm}|}{Physical Science}			& Physics, Atomic, Molecular \& Chemical &	17,010			\\ \hline

 			       									                & Geography &	3,908			\\ 
 			   \multicolumn{1}{|L{3cm}|}{Earth Science}				& Geology &	2,153 \\
 			    														& Astronomy \& Astrophysics &	22,825 \\ \hline

 			   \multicolumn{1}{|L{3cm}|}{Formal Science}				& Mathematics, Applied &	27,982			\\ \hline

 			  \end{tabular}
	\label{table:sampleSt}
	\end{table}

% A sample with proportionate stratification is chosen such that the distribution of observations in each stratum of the sample is the same as the distribution of observations in each stratum within the population.

% Stratified random sampling is used when the researcher wants to highlight a specific subgroup within the population. This technique is useful in such researches because it ensures the presence of the key subgroup within the sample.

For both the corpus and the sample, two classes of descriptive statistics are further calculated: \textit{location statistics} (mean, median and quantiles) and \textit{dispersion statistics} (standard deviation and standard error). Statistics calculated are: 

\begin{itemize}
\item \textit{$N$:} Number of documents in the corresponding set.
\item \textit{Max:} Maximum citation counts in the corresponding set.
\item \textit{Min:} Minimum citation counts in the corresponding set.
\item \textit{Mean ($\mu$):} The arithmetic average of citation counts in the corresponding set. We note that as the mean is not a robust measure of central tendency, it will be very sensitive to even one aberrant value (very high citation) out of $N$ documents. 
\item \textit{$Q_{1}$:} The first quartile. It indicates the median of the lower half of the data. This means that 25\% of documents cited less than $Q_{1}$.
\item \textit{$Q_{2}$ (median):} The middle number in the data. The median is a robust statistic against an outlier and can resist up to 50 \% of outliers. 
\item \textit{$Q_{3}$:} The third quartile. It indicates the median of the upper half of the data, which means that 75\% of documents cited less than $Q_{3}$.
\item \textit{$\sigma$:} Standard deviation. It is the dispersion of the data. It shows how accurately the mean represents the sample. 
\item \textit{$SE$:} Estimated standard error of the mean. It measures how far the sample mean  is likely to be from the population mean. That is, it is a measure of how precise our estimate of the mean is. 
\end{itemize}

Descriptive statistics for the corpus are presented in Table \ref{table:decStats}. The average citation in the corpus is approximately 9 and 29\% of the documents are cited more than 9 times. Although the maximum number of citation is 8,234, from the values of the upper and lower quartiles we can conclude that approximately 50\% of the documents are cited between 1-9 times in the corpus. The statistics for the sample can also be found in the table. The mean value and the quartiles are very similar to those for the corpus, but with a lower maximum citation count.

\begin{table}[bth]
\centering
\caption{Descriptive statistics for times cited in the corpus and samples}
\renewcommand\arraystretch{1}
\scriptsize
\begin{tabular}	{L{3.4cm} R{1.2cm} R{.7cm} R{.6cm} R{.7cm}  R{.4cm}  R{.4cm} R{.4cm}  R{.7cm} R{.7cm}} 
\hline

\multicolumn{1}{C{3.4cm}}{\textbf{Set}} & \multicolumn{1}{C{1.2cm}}{\textbf{$N$}} & \multicolumn{1}{C{.7cm}}{\textbf{Max}} &\multicolumn{1}{C{.6cm}}{\textbf{Min}}  & \multicolumn{1}{C{.7cm}}{\textbf{$\mu$}}  & \multicolumn{1}{C{.4cm}}{\textbf{$Q_{1}$}} &\multicolumn{1}{C{.4cm}}{\textbf{$Q_{2}$}} & \multicolumn{1}{C{.4cm}}{\textbf{$Q_{3}$}} & \multicolumn{1}{C{,7cm}}{\textbf{$\sigma$}} &\multicolumn{1}{C{.7cm}}{\textbf{SE}} \\ \hline

Corpus 	&	1,673,350	&8,234	&	0	&	9.38	&1&4&11 & 23.33& 0.018 \\ 
Sample 	&	147,901	&1,508	&	0	&	9.43	&1&5&11 & 18.30& 0.047 \\ \hline
$Sample_{1}$ (Life Science)	&	29,661	&978	&	0	&	10.95	&2&6&13 &19.96& 0.116 \\ 
$Sample_{2}$ (Social Science)	&	31,159	&304	&	0	&	7.70	&1&4&10 & 12.83& 0.073 \\ 
$Sample_{3}$ (Physical Science)	&	30,213	&1,508	&	0	&	11.57	&2&7&14 & 21.86& 0.126 \\ 
$Sample_{4}$ (Earth Science)	&	28,886	&1,243	&	0	&	12.44	&2&7&15 & 22.81& 0.134 \\ 
$Sample_{5} $(Formal Science)	&	27,982	&	253	&	0	&	4.33	&0&2&5& 8.03	& 0.048 \\ \hline
Biology &	9,917	&	517	&	0	&	10.36	& 2&6&13&16.09&0.162    \\
Medicine, Research \& Experimental &	19,744	&	978	&	0	&	11.24	&  2&6&13&21.63&0.154 \\ \hline

Psychology  &  6,989	&	300	&	0	&	10.56	&	3	&	7	&	14	&	14.75	&	0.176  \\
Political Science &	5,106	&	293	&	0	&	7.04	&	1	&	4	&	9	&	10.89	&	0.152 \\
Management &	14,339	&	304	&	0	&	6.92	&	0	&	2	&	8	&	13.30	&	0.111 \\
Sociology &	4,725	&	192	&	0	&	6.54	&	1	&	4	&	8	&	9.17	&	0.133 \\ \hline

Physics, Particles \& Fields  &	13,203	&	1,508	&	0	&	11.72	&	2	&	6	&	14	&	22.81	&	0.199 \\
Physics, Atomic, Molecular \& Chemical &	17,010	&	1,143	&	0	&	11.46	&	3	&	7	&	14	&	21.09	&	0.162 \\ \hline

Geography &	3,908	&	775	&	0	&	10.27	&	2	&	6	&	13	&	18.95	&	0.303  \\
Geology &	2,153	&	108	&	0	&	7.86	&	2	&	5	&	10	&	9.18	&	0.198 \\
Astronomy \& Astrophysics &	22,825	&	1,243	&	0	&	13.24	&	2	&	7	&	16	&	24.20	&	0.160  \\ \hline

Mathematics, Applied &	27,982	&	253	&	0	&	4.33	&0&2&5& 8.03	& 0.048  \\ \hline

 \end{tabular}
\label{table:decStats}
\end{table}

\begin{figure}[htb]
  \centering
  \includegraphics[height=.22\textheight, width=.5\textwidth]{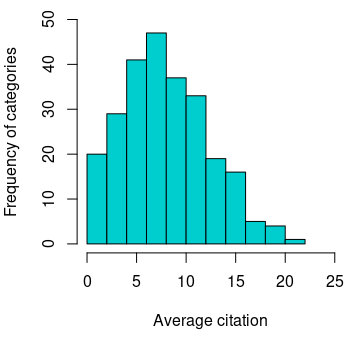}
  \caption{Frequency of categories versus average citation counts}
  \label{fig:NoCatAvCit}
\end{figure}

In the LSC, the average citation counts in categories vary. Descriptive statistics for each category are presented in Table \ref{DesStatsCats} and the Figure \ref{fig:NoCatAvCit}. As can be seen from the histogram, for more than the half of categories the average citation lies between 5 and 10. Almost 20 categories have documents cited 2 times on average. We suspect that research fields and the number of documents assigned to categories may be two factors influencing low citation counts. The scope of the category may be a factor that affects the citation counts. Documents in multidisciplinary categories may be cited by researchers from different fields. For instance, the maximum citation in the category `Multidisciplinary Sciences' is 3,004 with a relatively high mean 18.32, first quartile 4 and third quartile 18. This indicated that 50\% of documents are cited 4-18 times, which is high relative to the corpus.

In Table \ref{table:MaxCit} the ten categories with the highest citation counts in the corpus are tabulated with the first, second and third quartiles. We can see that categories involving several academic disciplines appear in this list. It should be noted that the category `Computer Science, Interdisciplinary Applications' has relative small quartiles when compared with the other categories. This means that citation counts for this category are low in general even if there exist highly cited documents. This suggests that this fact may be characteristic for some categories. Therefore, it is also important to look at categories with almost 0 citation. We list categories with very low third quartile value in Table \ref{table:LowCit}. It can be clearly seen that papers in discourse studies and fine arts are cited a few times in general. This is also true for computer science and engineering categories that have a high number of documents.

\begin{table}[bth]
\centering
\caption{Categories with the maximum citation counts in the corpus}
\renewcommand\arraystretch{1}	
\scriptsize
\begin{tabular}	{|L{6.5cm}| R{1cm} |R{1cm}| R{1cm}| R{1cm}|}
\hline
  	  \multicolumn{1}{|L{6.5cm}|}{\textbf{Category}} & \multicolumn{1}{C{1cm}|}{\textbf{Max}} & \multicolumn{1}{C{1cm}|}{\textbf{$Q_{1}$}}& \multicolumn{1}{C{1cm}|}{\textbf{$Q_{2}$}}& \multicolumn{1}{C{1cm}|}{\textbf{$Q_{3}$}} \\ \hline

Oncology	                 &		8,234	&			4	&	9	&	17 \\ \hline
Genetics \& Heredity		&	6,756		&	4	&	8	&	17 \\ \hline
Computer Science, Interdisciplinary Applications		&	4,744	&	0	&	1	&	6  \\ \hline
Biotechnology \& Applied Microbiology		&	4,744	&		3	&	7	&	14  \\ \hline
Biochemical Research Methods		&	4,744	&		3	&	8	&	14.75  \\ \hline
Statistics \& Probability		&	4,744	&		1	&	3	&	7  \\ \hline
Mathematical \& Computational Biology		&	4,744	&		1	&	4	&	9  \\ \hline
Medicine, General \& Internal		&	3,908	&		2	&	4	&	10  \\ \hline
Multidisciplinary Sciences		&	3,004	&		4	&	9	&	18  \\ \hline
Materials Science, Multidisciplinary		&	2,464	&		0	&	4	&	12  \\ \hline
Nanoscience \& Nanotechnology		&	2,464	&		2	&	8	&	20  \\ \hline

\end{tabular}
\label{table:MaxCit}
\end{table}

\begin{table}[bth]
\centering
\caption{Categories with low citation counts in the corpus}
\renewcommand\arraystretch{1}	
\scriptsize
\begin{tabular}	{|L{5cm}| R{1cm} |R{1cm} | R{1cm}| R{1cm}| R{1cm}|}
\hline
  	  \multicolumn{1}{|L{5cm}|}{\textbf{Category}} & \multicolumn{1}{C{1cm}|}{\textbf{$N$}} & \multicolumn{1}{C{1cm}|}{\textbf{Max}} & \multicolumn{1}{C{1cm}|}{\textbf{$Q_{1}$}}& \multicolumn{1}{C{1cm}|}{\textbf{$Q_{2}$}}& \multicolumn{1}{C{1cm}|}{\textbf{$Q_{3}$}} \\ \hline

Literary Theory \& Criticism	&	498	&	45	&	0	&	0	&	0  \\ \hline
Dance	&	74	&	8		&	0	&	0	&	0  \\ \hline
Literature, Slavic	&	35	&	3	&	0	&	0	&	0  \\ \hline
Literary Reviews	&	35	&	2	&		0	&	0	&	0  \\ \hline
Architecture	&	1,376	&	145	&		0	&	0	&	1  \\ \hline
Humanities, Multidisciplinary	&	2,559	&	53	&		0	&	0	&	1  \\ \hline
Asian Studies	&	877	&	32		&	0	&	0	&	1  \\ \hline
Literature	&	1,608	&	18		&	0	&	0	&	1  \\ \hline
Medieval \& Renaissance Studies	&	485	&	11	&		0	&	0	&	1  \\ \hline
Classics	&	325	&	9		&	0	&	0	&	1  \\ \hline

Engineering, Electrical \& Electronic	&	174,272	&	2,028		&	0	&	0	&	3 \\ \hline
Telecommunications	&	40,550	&	2,028	&		0	&	1	&	3 \\ \hline
Computer Science, Information Systems	&	45,865	&	715	&		0	&	1	&	3 \\ \hline
Computer Science, Hardware \& Architecture	&	18,489	&	532	&		0	&	1	&	3 \\ \hline
Language \& Linguistics	&	5,174	&	112	&		0	&	1	&	3 \\ \hline

\end{tabular}
\label{table:LowCit}
\end{table}

\begin{figure}[htb]
  \centering
  \includegraphics[height=.25\textheight, width=.85\textwidth]{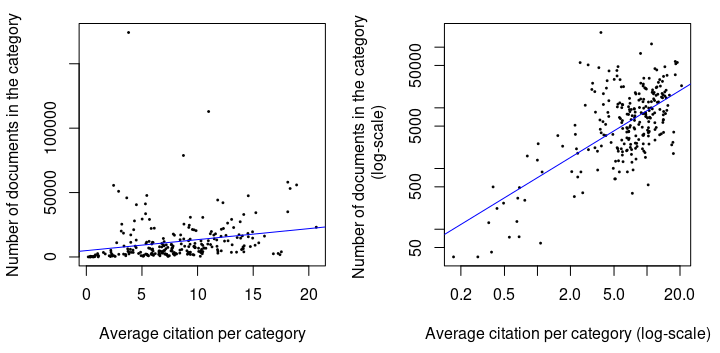}
  \caption{Number of documents per category versus the average citation counts in the category}
  \label{fig:1stPlot}
\end{figure}

\begin{figure}[htb]
  \centering
  \includegraphics[height=.25\textheight, width=.85\textwidth]{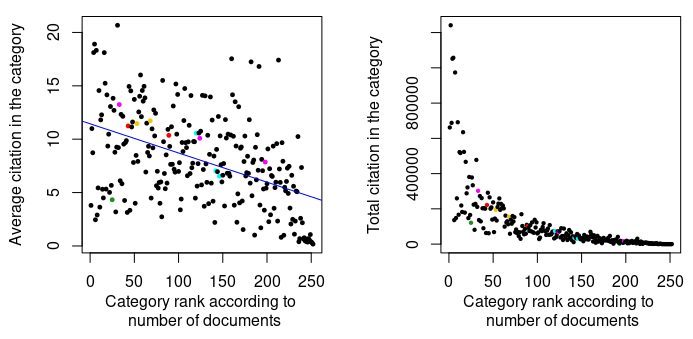}
  \caption{Average citation per categories versus category rank according to the number of documents. Colours indicate the categories in the sample: red is for life science categories, cyan is for social science categories, yellow is for physical science categories, pink is for earth science categories and green is for formal science category. }
  \label{fig:imageLine}
\end{figure}

In general, coverage of the natural sciences and medicine are much richer than for other branches in the LSC. However, we did not observe any explicit correlation between the average citation count and the number of documents. Figure \ref{fig:1stPlot} shows the number of documents per category versus the average citation counts in the category, with the linear regression line. We can see a slightly increasing trend with a wide spread of points around the line.  The figure also indicates that two categories with a high documents count can have very different citation counts with different means. For instance, the number of documents of `Engineering, Electrical \& Electronic' is 174,272 and `Materials Science, Multidisciplinary' is 112,912; however, the average citation count for `Materials Science, Multidisciplinary' is 3 times more than the average for `Engineering, Electrical \& Electronic' (Table \ref{table:topRank}). We can also see the same holds for quartiles. This fact is also confirmed in the Figure \ref{fig:imageLine} and Table \ref{table:topmean}. A slightly decreasing trend of average citation with the rank of categories is observable in the figure, but we can also see the spread of points. Two closely ranked categories can have very different statistics and two categories with very similar statistics can have very different rank in the corpus. Therefore, there may be other factors that contributes to citation counts, for instance, the field of research.

\begin{table}[bth]
\centering
\caption{Top ranked five categories (by the number of documents)}
\renewcommand\arraystretch{1}	
\scriptsize
\begin{tabular}	{|L{5cm}| R{1cm} |R{.8cm} | R{.8cm}| R{.7cm}| R{.5cm}| R{.5cm}|R{.5cm}|}
\hline
  	  \multicolumn{1}{|L{5cm}|}{\textbf{Category}} & \multicolumn{1}{C{1cm}|}{\textbf{$N$}} & \multicolumn{1}{C{.8cm}|}{\textbf{Max}} & \multicolumn{1}{C{.8cm}|}{\textbf{Min}} & \multicolumn{1}{C{.7cm}|}{\textbf{$\mu$}}& \multicolumn{1}{C{.5cm}|}{\textbf{$Q_{1}$}}& \multicolumn{1}{C{.5cm}|}{\textbf{$Q_{2}$}}& \multicolumn{1}{C{.5cm}|}{\textbf{$Q_{3}$}} \\ \hline

Engineering, Electrical \& Electronic	&	174,272	&	2,028	&	0	&	3.80	&	0	&	0	&	3  \\
Materials Science, Multidisciplinary	&	112,912	&	2,464	&	0	&	10.99	&	0	&	4	&	12\\
Physics, Applied	&	78,796	&	2,288	&	0	&	8.73	&	0	&	3	&	9\\
Chemistry, Physical	&	58,065	&	2,288	&	0	&	18.11	&	5	&	10	&	20\\
Chemistry, Multidisciplinary	&	55,907	&	2,210	&	0	&	18.90	&	3	&	9	&	21\\ \hline

\end{tabular}
\label{table:topRank}
\end{table}

\begin{table}[bth]
\centering
\caption{Categories with the highest average citation counts}
\renewcommand\arraystretch{1}	
\scriptsize
\begin{tabular}	{|L{5cm}| R{1cm} |R{.8cm} | R{.8cm}| R{.7cm}| R{.5cm}| R{.5cm}|R{.5cm}|}
\hline
  	  \multicolumn{1}{|L{5cm}|}{\textbf{Category}} & \multicolumn{1}{C{1cm}|}{\textbf{$N$}} & \multicolumn{1}{C{.8cm}|}{\textbf{Max}} & \multicolumn{1}{C{.8cm}|}{\textbf{Min}} & \multicolumn{1}{C{.7cm}|}{\textbf{$\mu$}}& \multicolumn{1}{C{.5cm}|}{\textbf{$Q_{1}$}}& \multicolumn{1}{C{.5cm}|}{\textbf{$Q_{2}$}}& \multicolumn{1}{C{.5cm}|}{\textbf{$Q_{3}$}} \\ \hline

Cell Biology	&	23,108	&	978	&	0	&	20.67	&	6	&	12	&	24	\\
Chemistry, Multidisciplinary	&	55,907	&	2,210	&	0	&	18.9	0&	3	&	9	&	21	\\
Multidisciplinary Sciences	&	53,140	&	3,004	&	0	&	18.32	&	4	&	9	&	18	\\
Chemistry, Physical	&	58,065	&	2,288	&	0	&	18.11	&	5	&	10	&	20	\\
Nanoscience \& Nanotechnology	&	35,050	&	2,464	&	0	&	18.11	&	2	&	8	&	20	\\
Critical Care Medicine	&	3,982	&	320	&	0	&	17.53	&	5	&	11	&	21	\\
Allergy	&	1,765	&	488	&	0	&	17.41	&	4	&	10	&	20	\\
Neuroimaging	&	2,702	&	527	&	0	&	17.24	&	6	&	12	&	22	\\
Cell \& Tissue Engineering	&	2,455	&	261	&	0	&	16.81	&	4	&	10	&	20	\\
Medicine, General \& Internal	&	16,179	&	3,908	&	0	&	16.01	&	2	&	4	&	10	\\ \hline

\end{tabular}
\label{table:topmean}
\end{table}

\section{Citation Classification} \label{CitClass}
In this section, we explore the efficacy of semantic meaning of scientific articles represented in the MS in predicting the impact of articles on basis of the LSC. We have employed classification algorithms in order to analyse the predictability of high/low citation of scientific papers. 

We apply the methods for binary classification and investigate the binary classification problem for differentiating between two types of impacts of scientific papers: little-cited (L) and highly-cited (H) papers. Our initial hypothesis is that, if the semantic meaning of texts indicate the impact of papers, a classifier should perform significantly better than chance (50\%). 

Distinguishing between highly-cited and little-cited papers can be done using various definitions. Two basic approaches to identify classes are to use absolute and relative thresholds \cite{aksnes}. Using an absolute number of citations for the entire corpus leads to the favouring of papers from highly-cited categories due to the differences in the average citation within disciplines. Instead, relative thresholds define the highly-cited and little-cited papers in each scientific category.

In this study, we begin with using relative thresholds for individual categories. Two rules to define thresholds for highly-cited and little-cited classes are used. In the first scheme, papers assigned to a category are divided into H/L classes according to the average citation in the category. A paper is labelled as highly-cited (H) if it has received more than the average citation, little-cited (L) otherwise. 

We then conduct an experiment to classify papers according to extremely highly-cited (EH) and extremely little-cited (EL) papers. In this scheme, papers in a certain sample are divided into four partitions by its corresponding quartiles from citations, where papers belong to one of the four partitions, described as: 
\begin{itemize}

\item $P_{1}$: papers cited less than or equal to the lower quartile $ Q_{1} $ 
\item $P_{2}$: papers cited more than $ Q_{1} $ and less than or equal to the median $ Q_{2} $
\item $P_{3}$: papers cited more than $ Q_{2} $ and less than the upper quartile $ Q_{3} $
\item $P_{4}$: papers cited more than or equal to $ Q_{4} $.
\end{itemize}  

In this experiment, we use articles from the set $P_{1}\cup P_{4}$. This implies that we performed binary classification on the subset of texts with citations more than $ Q_{3} $ and less than $ Q_{1} $. The partitions $P_{1}$ and $P_{4}$ are defined as the groups of two extreme citations where papers in the partitions $P_{1}$ and $P_{4}$ are labelled as extremely little-cited (EL) and extremely highly-cited (EH) respectively.

Two classification methods are initially investigated for comparison: Linear Discriminant Analysis (LDA) with Fisher’s discriminant rule and $k$-nearest neighbour (kNN). We follow procedures:

\begin{enumerate}
\item We apply Fisher's LDA in classifying scientific papers by considering the selected sample set of texts as a whole training set.
\item We apply kNN in classifying scientific papers by considering the selected sample set of texts as a whole training set.
\item We apply leave-one-out cross validation (LOOCV) in classifying scientific papers for the best LDA separation for demonstration of statistical power of the classifier.
\end{enumerate}

We finally extend the study to using supervised PCA defined for the classification problem. By supervised PCA, the dimension of FVTs will be reduced and the procedures for classification of papers defined above will be repeated with the reduced dimensionality. 

The evaluation of the potential impact of articles through the semantic analysis is made by using different representations of the meaning of texts. Vectors $FVT_{1}(d), FVT_{2}(d), FVT_{3}(d), FVT_{4}(d), FVT_{5}(d)$ in the constructed space are created and compared for classification.

We have selected three categories from three different branches of science in Table \ref{table:decStats} for the experimental study: Mathematics, Applied, Biology and Management. The numbers of texts assigned to each class in categories  are given in Table \ref{table:textNum}.

\begin{table*}\centering
\caption{Number of texts assigned to categories selected in each class}
\scriptsize
\begin{tabular}{@{}lrrrcrrr@{}}\toprule
&\multicolumn{3}{c}{H/L classes} &\phantom{ab}& \multicolumn{3}{c}{EH/EL classes}  \\
\cmidrule{2-4} \cmidrule{6-8}  
 Category & \multicolumn{1}{c}{Class H}& \multicolumn{1}{c}{Class L } &  \multicolumn{1}{c}{Total } &  \phantom{ab} &  \multicolumn{1}{c}{Class EH}& \multicolumn{1}{c}{Class EL } & \multicolumn{1}{c}{Total }    \\
\midrule
Mathematics, Applied	&	7,975	&	20,007	& 27,982&&	7,975	&7,925&	15,900	\\
Biology	 	&	3,055	&	6,862	&	9,917		&&	2,553	&	2,772	&	5,325		\\
Management	&	4,426	&	9,913	&	14,339		&&	3,980	& 	5,257 	&	9,237		\\
\bottomrule
\end{tabular}
\label{table:textNum}
\end{table*}

In kNN, to avoid the impact of the uneven distribution of sample size in classes, we tested kNN with weights as described in Section \ref{evaluation}. This modification achieved better classification performance for these imbalanced classes. Thus we only present results for weighted kNN. We applied kNN where the $k$ is 1, 3, 5, 7, 11, 13 and 17 for each vector. We have presented only the results for value of $k$ with the best classification performance in each case individually. 

\subsection{The spaces used in classification \nopunct}\hspace*{\fill} \\\\
In this research, classification methods are compared for different vector representations in 3 different vector spaces. 
 \subsubsection{\textbf{Original Space}} The original space refers to the Meaning Space that is defined in Section \ref{textRep}. In this space, each text is represented by the FVTs using ${\rm FVT}_1$ or ${\rm FVT}_3$, respectively of 252 or 504. This means that each text has at least dimension 252 in the Meaning Space. 
 \subsubsection{\textbf{13-Dimensional Reduced Basis}} As described in \cite{CompMeaning}, words can be represented using PC axes; the optimal number of PC was computed to be 13 by the PCN-CN. In order to carry out additional investigations on the basis of 13-dimensional word space, we represented words in 13-Dimensional PC space and then constructed the FVTs based on these vectors as described in Section~\ref{texRepModel}. In this case, each text is represented by at least 13-dimensional vectors, for instance, the mean of 13 dimensional coordinates. We compared the performance of the classifiers for each vector in the original space and space with reduced basis. 

 \subsubsection{\textbf{The space constructed after Supervised PCA}}
For binary classification, we followed the following procedure. For each vector representation of text:

\begin{enumerate}

\item Apply PCA to the data in the original space.
\item Find the number of components by Kaiser and Broken Stick rules to be used the initial number of components in supervised PCA.
\item Initialise the number of components (ncomp) as the minimum number found in previous step. We used the Broken Stick results.
\item Apply supervised PCA with the components from 1 to ncomp, and represent all texts on this new space.
\item Apply LDA with the data on reduced spaces and calculate the sum of sensitivity and specificity.
\item Take the maximum of the sum of sensitivity and specificity, and identify the number of components for this number. Report the sum of sensitivity and specificity to evaluate the performance of LDA classifier.
\item Apply kNN with the identified number of dimension by LDA. Calculate the sum of sensitivity and specificity to evaluate the performance of the kNN classifier.
\end{enumerate}
	 
The number of components (dimensions of FVT vectors) in the original space and the reduced space constructed after applying supervised PCA for categories Mathematics, Applied, Biology and Management are presented in Tables \ref{table:dim1Math}-\ref{table:dim3man}.

\begin{table*}\centering
\caption{Dimensions of the FVT vectors in the original space and the reduced space constructed after applying supervised PCA for the category Mathematics, Applied}
\scriptsize
\begin{tabular}{@{}lrrrrcrrr@{}}\toprule
&& \multicolumn{3}{c}{H/L classes} &\phantom{ab}& \multicolumn{3}{c}{EH/EL classes}  \\
\cmidrule{3-5} \cmidrule{7-9}  
 & \multicolumn{1}{c}{Original} & \multicolumn{1}{c}{Kaiser}& \multicolumn{1}{c}{Broken } &\multicolumn{1}{c}{Supervised} &  \phantom{ab} &  \multicolumn{1}{c}{Kaiser}& \multicolumn{1}{c}{Broken } &\multicolumn{1}{c}{Supervised}  \\
FVT & \multicolumn{1}{c}{Space} & \multicolumn{1}{c}{Rule}& \multicolumn{1}{c}{Stick} &\multicolumn{1}{c}{PCA} &  \phantom{ab} &  \multicolumn{1}{c}{Rule}& \multicolumn{1}{c}{Stick} &\multicolumn{1}{c}{PCA}\\
\midrule
$(\overrightarrow{\mu})$	&	252	&	36	&	15	&	14	&&	32	&	15	&	14	\\
$(\overrightarrow{PC_{1}})$	&	252	&	25	&	8	&	8	&&	24	&	8	&	8	\\
$(\overrightarrow{\mu} ,\overrightarrow{PC_{1}})$	&	504	&	59	&	22	&	16	&&	59	&	22	&	22	\\
$(\overrightarrow{c_{1}},\overrightarrow{c_{2}})$	&	504	&	41	&	20	&	19	&&	41	&	20	&	14	\\
$(\overrightarrow{c_{1}},\overrightarrow{c_{2}},\overrightarrow{PC_{1}})$	&	756	&	62	&	24	&	23	&&	63	&	24	&	24	\\
\bottomrule
\end{tabular}
\label{table:dim1Math}
\end{table*}

\begin{table*}\centering
\caption{Dimensions of FVT vectors in the original space and the space constructed after applying supervised PCA for the category Biology}
\scriptsize
\begin{tabular}{@{}lrrrrcrrr@{}}\toprule
&& \multicolumn{3}{c}{H/L classes} &\phantom{ab}& \multicolumn{3}{c}{EH/EL classes}  \\

\cmidrule{3-5} \cmidrule{7-9}  
 & \multicolumn{1}{c}{Original} & \multicolumn{1}{c}{Kaiser}& \multicolumn{1}{c}{Broken } &\multicolumn{1}{c}{Supervised} &  \phantom{ab} &  \multicolumn{1}{c}{Kaiser}& \multicolumn{1}{c}{Broken } &\multicolumn{1}{c}{Supervised}  \\

FVT & \multicolumn{1}{c}{Space} & \multicolumn{1}{c}{Rule}& \multicolumn{1}{c}{Stick} &\multicolumn{1}{c}{PCA} &  \phantom{ab} &  \multicolumn{1}{c}{Rule}& \multicolumn{1}{c}{Stick} &\multicolumn{1}{c}{PCA}\\

\midrule

$(\overrightarrow{\mu})$	&	252	&	35	&	13	&	12	&&	36	&	14	&	12	\\
$(\overrightarrow{PC_{1}})$	&	252	&	29	&	10	&	8	&&	30	&	11	&	9	\\
$(\overrightarrow{\mu} ,\overrightarrow{PC_{1}})$	&	504	&	64	&	25	&	25	&&	64	&	25	&	25	\\
$(\overrightarrow{c_{1}},\overrightarrow{c_{2}})$	&	504	&	50	&	23	&	23	&&	49	&	22	&	20	\\
$(\overrightarrow{c_{1}},\overrightarrow{c_{2}},\overrightarrow{PC_{1}})$	&	756	&	76	&	30	&	30	&&	76	&	30	&	30	\\

\bottomrule
\end{tabular}
\label{table:dim2bio}
\end{table*}

\begin{table*}\centering
\caption{Dimensions of FVT vectors in the original space and the reduced space constructed after applying supervised PCA for the category Management}
\scriptsize
\begin{tabular}{@{}lrrrrcrrr@{}}\toprule
&& \multicolumn{3}{c}{H/L classes} &\phantom{ab}& \multicolumn{3}{c}{EH/EL classes}  \\

\cmidrule{3-5} \cmidrule{7-9} 
 & \multicolumn{1}{c}{Original} & \multicolumn{1}{c}{Kaiser}& \multicolumn{1}{c}{Broken } &\multicolumn{1}{c}{Supervised} &  \phantom{ab} &  \multicolumn{1}{c}{Kaiser}& \multicolumn{1}{c}{Broken } &\multicolumn{1}{c}{Supervised}  \\

FVT & \multicolumn{1}{c}{Space} & \multicolumn{1}{c}{Rule}& \multicolumn{1}{c}{Stick} &\multicolumn{1}{c}{PCA} &  \phantom{ab} &  \multicolumn{1}{c}{Rule}& \multicolumn{1}{c}{Stick} &\multicolumn{1}{c}{PCA}\\

\midrule

$(\overrightarrow{\mu})$	&	252	&	36	&	13	&	11	&&	36	&	13	&	13	\\
$(\overrightarrow{PC_{1}})$	&	252	&	23	&	9	&	9	&&	24	&	9	&	9	\\
$(\overrightarrow{\mu} ,\overrightarrow{PC_{1}})$	&	504	&	62	&	23	&	23	&&	61	&	22	&	22	\\
$(\overrightarrow{c_{1}},\overrightarrow{c_{2}})$	&	504	&	39	&	16	&	11	&&	40	&	16	&	11	\\
$(\overrightarrow{c_{1}},\overrightarrow{c_{2}},\overrightarrow{PC_{1}})$	&	756	&	61	&	25	&	25	&&	62	&	25	&	25	\\

\bottomrule
\end{tabular}
\label{table:dim3man}
\end{table*}

\subsection{Results \nopunct}\hspace*{\fill} \\\\
Two classification algorithms, LDA and kNN, were executed to show how much information the semantics of texts have on impact of articles. We selected three categories to demonstrate the results of classification and the differences in efficiency of semantics on bibliometric characteristics of articles in different categories. 

\subsubsection{\textbf{Applied Mathematics}\nopunct}\hspace*{\fill} \\\\
The citation count is determined by the nature of the individual categories. There are many factors that contribute to the citation counts in individual research fields. Factors such as the size of the field, annual scientific production, and relevance of article for the research activities in a field are very influential on the number of citations.  

The category Applied Mathematics is selected as a sample from Formal Science - one of the five  main branches of science. It is noteworthy that articles assigned to the category Applied Mathematics usually address issues combining mathematical methods and specialised knowledge in specific fields such as physics, biology or business. Such categories generally suggest that there is no recurrent common characteristics of articles and there are large differences in citation patterns of papers from different disciplines. In Applied Mathematics, the separation of papers into highly popular fields may not be easily done. Therefore, the prediction of citation by semantics of such papers may be difficult for categories of multidisciplinary publications. 

For Applied Mathematics, classification results for our two algorithms in the different spaces are shown in Tables \ref{table:mathsRes1}-\ref{table:mathsRes4}. The first column in the tables gives the type of FVT to represent the text. The other columns present the sensitivity (\%), specificity (\%) and the sum of sensitivity and specificity (\%) in each experiment. Sensitivity is the proportion of correctly identified highly-cited papers and specificity is the proportion of correctly identified little-cited papers.  The results of LDA and kNN are recorded in separate tables. For kNN (weighted kNN), results are given only for $k$ with the best classification performance, that is, the sum of sensitivity and specificity is the maximum. In each table, we used red to highlight the case where the algorithm has achieved the highest classification performance according to the criterion used. All procedures are repeated for H/L and EH/EL classification.

For H/L classification, results for LDA on three spaces are shown in the Table \ref{table:mathsRes1}. We can see that LDA performs best in the Original Space for all the vector representations with sensitivity 64.15\%, specificity 58.85\% and the sum 123\%. The next best results are obtained in the space constructed by supervised PCA with the 13-dimensional reduced basis. We can also see that the vector  $(\overrightarrow{c_{1}},\overrightarrow{c_{2}},\overrightarrow{PC_{1}})$ performs better than the others. $(\overrightarrow{\mu} ,\overrightarrow{PC_{1}})$ achieves the second best performance in all cases. In general, we can conclude that adding $\overrightarrow{PC_{1}}$ to vectors $\overrightarrow{\mu} $ and $(\overrightarrow{c_{1}},\overrightarrow{c_{2}})$  improves classification performance.

\begin{table*}\centering
\caption{Results of the citation classifier LDA according to H/L for the category Mathematics, Applied}
\scriptsize
\begin{tabular}{@{}lrrrcrrrcrrr@{}}\toprule
 & \multicolumn{3}{c}{Original Space} & \phantom{ab}& \multicolumn{3}{c}{13-Dimensional } &
\phantom{ab} & \multicolumn{3}{c}{Supervised PCA}\\

 & \multicolumn{3}{c}{} & \phantom{ab}& \multicolumn{3}{c}{Reduced Basis} &
\phantom{ab} & \multicolumn{3}{c}{}\\
\cmidrule{2-4} \cmidrule{6-8} \cmidrule{10-12}

& $Sens.$ &  $Spec.$ &  $Sum$  &&  $Sens.$ & $Spec.$ &  $Sum$ &&  $Sens.$ &  $Spec.$ &  $Sum$\\ 

FVT  & $(\%)$ &  $(\%)$ &  $(\%)$  &&  $(\%)$ & $(\%)$ &  $(\%)$ &&  $(\%)$ &  $(\%)$ &  $(\%)$\\ 
\midrule
$(\overrightarrow{\mu})$	&	67.91	&	49.61	&	117.52	&&	77.02	&	29.14	&	106.16	&&	68.28	&	44.55	&	112.83	\\
$(\overrightarrow{PC_{1}})$	&	61.66	&	57.02	&	118.68	&&	63.42	&	44.68	&	108.10	&&	63.49	&	47.24	&	110.73	\\
$(\overrightarrow{\mu} ,\overrightarrow{PC_{1}})$	&	58.83	&	63.73	&	122.56	&&	62.91	&	46.47	&	109.38	&&	59.12	&	54.37	&	113.49	\\
$(\overrightarrow{c_{1}},\overrightarrow{c_{2}})$	&	67.40	&	49.53	&	116.93	&&	49.54	&	58.89	&	108.43	&&	57.93	&	50.82	&	108.75	\\
$(\overrightarrow{c_{1}},\overrightarrow{c_{2}},\overrightarrow{PC_{1}})$	&	64.15	&	58.85	&	\hlc[pink]{123.00}	&&	55.37	&	54.96	&	110.33	&&	54.13	&	59.47	&	113.61	\\

\bottomrule
\end{tabular}
\label{table:mathsRes1}
\end{table*} 

The binary classification for classes EH/EL achieves higher performance results for all experiments. This implies that LDA performs better for distinguishing the two extreme citation classes. These classes also have higher sensitivity and specificity values, showing that the semantics in papers with two extreme citation counts are clearly different. The best performance is still achieved when using the $(\overrightarrow{c_{1}},\overrightarrow{c_{2}},\overrightarrow{PC_{1}})$ in the original space, where sensitivity 67.25\%, specificity 62.97\% and the sum 130.21\%. In general, the results show that the performance of LDA has similar behaviour in all spaces.

\begin{table*}\centering
%\ra{1.3}
\caption{Results of the citation classifier LDA according to EH/EL for the category Mathematics, Applied}
\scriptsize
\begin{tabular}{@{}lrrrcrrrcrrr@{}}\toprule
 & \multicolumn{3}{c}{Original Space} & \phantom{ab}& \multicolumn{3}{c}{13-Dimensional } &
\phantom{ab} & \multicolumn{3}{c}{Supervised PCA}\\

 & \multicolumn{3}{c}{} & \phantom{ab}& \multicolumn{3}{c}{Reduced Basis} &
\phantom{ab} & \multicolumn{3}{c}{}\\
\cmidrule{2-4} \cmidrule{6-8} \cmidrule{10-12}

& $Sens.$ &  $Spec.$ &  $Sum$  &&  $Sens.$ & $Spec.$ &  $Sum$ &&  $Sens.$ &  $Spec.$ &  $Sum$\\ 

FVT  & $(\%)$ &  $(\%)$ &  $(\%)$  &&  $(\%)$ & $(\%)$ &  $(\%)$ &&  $(\%)$ &  $(\%)$ &  $(\%)$\\ 
\midrule

$(\overrightarrow{\mu})$	&	69.93	&	54.69	&	124.62	&&	74.37	&	35.51	&	109.88	&&	69.93	&	47.31	&	117.24	\\
$(\overrightarrow{PC_{1}})$	&	65.20	&	58.90	&	124.11	&&	69.82	&	39.70	&	109.52	&&	66.68	&	47.34	&	114.03	\\
$(\overrightarrow{\mu} ,\overrightarrow{PC_{1}})$	&	71.56	&	58.37	&	129.93	&&	68.60	&	44.69	&	113.30	&&	65.08	&	52.61	&	117.68	\\
$(\overrightarrow{c_{1}},\overrightarrow{c_{2}})$	&	68.66	&	54.16	&	122.82	&&	55.94	&	55.04	&	110.98	&&	53.98	&	56.76	&	110.74	\\
$(\overrightarrow{c_{1}},\overrightarrow{c_{2}},\overrightarrow{PC_{1}})$	&	67.25	&	62.97	&	\hlc[pink]{130.21}	&&	54.19	&	58.03	&	112.23	&&	64.89	&	53.56	&	118.45	\\

\bottomrule
\end{tabular}
\label{table:mathsRes2}
\end{table*}

In order to carry out additional investigation on the basis of the findings from LDA, we performed LOOCV in H/L classification by LDA in the Original Space only and compared the results (Table \ref{table:mathLOOCV}). As the performance in each experiment and general trends and values with and without LOOCV are very similar, we did not present LDA with LOOCV for other spaces and extreme citation classification. In this case, the best performance is achieved for the vector $(\overrightarrow{\mu} ,\overrightarrow{PC_{1}})$, only marginally better than from the vector $(\overrightarrow{c_{1}},\overrightarrow{c_{2}},\overrightarrow{PC_{1}})$.

\begin{table*}\centering
\caption{Results of the citation classifier LDA with LOOCV in according to H/L for the category Mathematics, Applied (for the Original Space only)}
\scriptsize
\begin{tabular}{@{}lrrr@{}}\toprule
 & $Sens.$ &  $Spec.$ &  $Sum$  \\ 
FVT  & $(\%)$ &  $(\%)$ &  $(\%)$  \\ 
\midrule
$(\overrightarrow{\mu})$	&	65.24	&	48.29	&	113.53	\\
$(\overrightarrow{PC_{1}})$	&	58.36	&	56.35	&	114.71	\\
$(\overrightarrow{\mu} ,\overrightarrow{PC_{1}})$	&	60.30	&	57.90	&	\hlc[pink]{118.20}	\\
$(\overrightarrow{c_{1}},\overrightarrow{c_{2}})$	&	62.87	&	48.25	&	111.12	\\
$(\overrightarrow{c_{1}},\overrightarrow{c_{2}},\overrightarrow{PC_{1}})$	&	59.00	&	56.00	&	115.5	\\

\bottomrule
\end{tabular}
\label{table:mathLOOCV}
\end{table*}

Tables \ref{table:mathsRes3} and \ref{table:mathsRes4} present results for the kNN classifier for classes H/L and EH/EL, respectively. Similar to LDA, we observed that kNN achieved the highest sum corresponds to the extreme citation classification (EH/EL) with the vector $(\overrightarrow{\mu})$ in the space constructed by supervised PCA. The sensitivity reaches to 70.53\% and the specificity is 45.94\% with a sum of 116.48\% in this case. However, there is only a slight difference between values in this space and the Original Space.

\begin{table*}\centering
%\ra{1.3}
\caption{The best results for the kNN citation classifier for H/L classification for each FVT for the category Mathematics, Applied}
\scriptsize
\begin{tabular}{@{}lrrrrcrrrrcrrrr@{}}\toprule
 & \multicolumn{4}{c}{Original Space} & \phantom{a}& \multicolumn{4}{c}{13-Dimensional } &
\phantom{a} & \multicolumn{4}{c}{Supervised PCA}\\

 & \multicolumn{4}{c}{} & \phantom{a}& \multicolumn{4}{c}{Reduced Basis} &
\phantom{a} & \multicolumn{4}{c}{}\\
\cmidrule{2-5} \cmidrule{7-10} \cmidrule{12-15}

&$k$& $Sens.$ &  $Spec.$ &  $Sum$  &&  $k$& $Sens.$ & $Spec.$ &  $Sum$ && $k$&  $Sens.$ &  $Spec.$ &  $Sum$\\ 

FVT  && $(\%)$ &  $(\%)$ &  $(\%)$  && & $(\%)$ & $(\%)$ &  $(\%)$ && & $(\%)$ &  $(\%)$ &  $(\%)$\\ 
\midrule

$(\overrightarrow{\mu})$	&	11	&	62.60	&	48.16	&	\hlc[pink]{110.76}	&&	13	&	65.53	&	41.97	&	107.50	&&	17	&	68.14	&	41.06	&	109.19	\\
$(\overrightarrow{PC_{1}})$	&	13	&	68.46	&	40.76	&	109.22	&&	17	&	63.54	&	44.16	&	107.70	&&	17	&	62.82	&	44.13	&	106.95	\\
$(\overrightarrow{\mu} ,\overrightarrow{PC_{1}})$	&	13	&	68.46	&	40.77	&	109.23	&&	11	&	52.87	&	55.75	&	108.62	&&	17	&	65.54	&	43.10	&	108.64	\\
$(\overrightarrow{c_{1}},\overrightarrow{c_{2}})$	&	11	&	57.30	&	50.07	&	107.38	&&	13	&	63.50	&	43.06	&	106.56	&&	13	&	61.87	&	43.68	&	105.55	\\
$(\overrightarrow{c_{1}},\overrightarrow{c_{2}},\overrightarrow{PC_{1}})$	&	11	&	56.25	&	53.20	&	109.45	&&	17	&	66.78	&	41.02	&	107.80	&&	17	&	65.96	&	42.77	&	108.73	\\

\bottomrule
\end{tabular}
\label{table:mathsRes3}
\end{table*}

\begin{table*}\centering
%\ra{1.3}
\caption{The best results for the kNN citation classifier for EH/EL classification for each FVT for the category Mathematics, Applied}
\scriptsize
\begin{tabular}{@{}lrrrrcrrrrcrrrr@{}}\toprule
 & \multicolumn{4}{c}{Original Space} & \phantom{a}& \multicolumn{4}{c}{13-Dimensional } &
\phantom{a} & \multicolumn{4}{c}{Supervised PCA}\\

 & \multicolumn{4}{c}{} & \phantom{a}& \multicolumn{4}{c}{Reduced Basis} &
\phantom{a} & \multicolumn{4}{c}{}\\
\cmidrule{2-5} \cmidrule{7-10} \cmidrule{12-15}

&$k$& $Sens.$ &  $Spec.$ &  $Sum$  &&  $k$& $Sens.$ & $Spec.$ &  $Sum$ && $k$&  $Sens.$ &  $Spec.$ &  $Sum$\\ 

FVT  && $(\%)$ &  $(\%)$ &  $(\%)$  && & $(\%)$ & $(\%)$ &  $(\%)$ && & $(\%)$ &  $(\%)$ &  $(\%)$\\ 
\midrule

$(\overrightarrow{\mu})$	&	11	&	77.83	&	37.99	&	115.82	&&	11	&	68.41	&	44.57	&	112.98	&&	17	&	70.53	&	45.94	&	\hlc[pink]{116.48}	\\
$(\overrightarrow{PC_{1}})$	&	17	&	73.52	&	41.22	&	114.74	&&	17	&	61.69	&	49.09	&	110.78	&&	17	&	61.87	&	49.55	&	111.42	\\
$(\overrightarrow{\mu} ,\overrightarrow{PC_{1}})$	&	17	&	73.52	&	41.22	&	114.74	&&	11	&	66.55	&	45.88	&	112.43	&&	17	&	67.74	&	46.97	&	114.70	\\
$(\overrightarrow{c_{1}},\overrightarrow{c_{2}})$	&	13	&	72.21	&	39.80	&	112.01	&&	17	&	65.92	&	44.40	&	110.32	&&	17	&	60.13	&	47.77	&	107.90	\\
$(\overrightarrow{c_{1}},\overrightarrow{c_{2}},\overrightarrow{PC_{1}})$	&	17	&	73.63	&	40.93	&	114.56	&&	17	&	66.22	&	45.31	&	111.53	&&	17	&	67.81	&	46.37	&	114.18	\\

\bottomrule
\end{tabular}
\label{table:mathsRes4}
\end{table*}

For Applied Mathematics, the best results with the sum of sensitivity and specificity being greater than 130\% were achieved for the vector $(\overrightarrow{c_{1}},\overrightarrow{c_{2}},\overrightarrow{PC_{1}})$ in the original space by the classifier LDA. This result corresponds to the classification of extreme citations (EH/EL) with 67.25\% sensitivity and 62.97\% specificity. This means that LDA with selection of the threshold by the sum of sensitivity and specificity maximisation is the best for classifying extreme citations.

\subsubsection{\textbf{Biology}\nopunct}\hspace*{\fill} \\\\
We selected the category Biology as a sample from the Life Science branch (see Table \ref{table:sampleSt}). The Tables \ref{table:bioRes1}-\ref{table:bioRes4} show results very similar to those obtained for Applied Mathematics. The classification performance is better for the separation of both H/L and EH/EL classes than we achieved for Applied Mathematics. 

In almost all experiments, results for both LDA and kNN show that classifiers perform better in the Original Space for all the FVTs, excluding the vector $(\overrightarrow{\mu} ,\overrightarrow{PC_{1}})$ when kNN is employed.  

\begin{table*}\centering

%\ra{1.3}
\caption{Results of the citation classifier LDA for H/L classification for the category Biology}
\scriptsize
\begin{tabular}{@{}lrrrcrrrcrrr@{}}\toprule
 & \multicolumn{3}{c}{Original Space} & \phantom{ab}& \multicolumn{3}{c}{13-Dimensional } &
\phantom{ab} & \multicolumn{3}{c}{Supervised PCA}\\

 & \multicolumn{3}{c}{} & \phantom{ab}& \multicolumn{3}{c}{Reduced Basis} &
\phantom{ab} & \multicolumn{3}{c}{}\\
\cmidrule{2-4} \cmidrule{6-8} \cmidrule{10-12}

& $Sens.$ &  $Spec.$ &  $Sum$  &&  $Sens.$ & $Spec.$ &  $Sum$ &&  $Sens.$ &  $Spec.$ &  $Sum$\\ 

FVT  & $(\%)$ &  $(\%)$ &  $(\%)$  &&  $(\%)$ & $(\%)$ &  $(\%)$ &&  $(\%)$ &  $(\%)$ &  $(\%)$\\ 
\midrule

$(\overrightarrow{\mu})$	&	74.30	&	69.21	&	143.51	&&	64.29	&	63.48	&	127.77	&&	70.47	&	64.33	&	134.80	\\
$(\overrightarrow{PC_{1}})$	&	72.50	&	63.87	&	136.38	&&	64.29	&	52.04	&	116.33	&&	71.72	&	55.80	&	127.52	\\
$(\overrightarrow{\mu} ,\overrightarrow{PC_{1}})$	&	75.38	&	71.60	&	\hlc[pink]{146.98}	&&	65.66	&	63.45	&	129.11	&&	70.77	&	59.82	&	130.59	\\
$(\overrightarrow{c_{1}},\overrightarrow{c_{2}})$	&	70.70	&	69.12	&	139.82	&&	67.53	&	50.45	&	117.98	&&	58.89	&	58.90	&	117.79	\\
$(\overrightarrow{c_{1}},\overrightarrow{c_{2}},\overrightarrow{PC_{1}})$	&	78.69	&	67.06	&	145.76	&&	68.02	&	55.09	&	123.11	&&	67.66	&	64.16	&	131.82	\\

\bottomrule
\end{tabular}
\label{table:bioRes1}
\end{table*}

\begin{table*}\centering
%\ra{1.3}
\caption{Results of the citation classifier LDA for EH/EL classification for the category Biology}
\scriptsize
\begin{tabular}{@{}lrrrcrrrcrrr@{}}\toprule
 & \multicolumn{3}{c}{Original Space} & \phantom{ab}& \multicolumn{3}{c}{13-Dimensional } &
\phantom{ab} & \multicolumn{3}{c}{Supervised PCA}\\

 & \multicolumn{3}{c}{} & \phantom{ab}& \multicolumn{3}{c}{Reduced Basis} &
\phantom{ab} & \multicolumn{3}{c}{}\\
\cmidrule{2-4} \cmidrule{6-8} \cmidrule{10-12}

& $Sens.$ &  $Spec.$ &  $Sum$  &&  $Sens.$ & $Spec.$ &  $Sum$ &&  $Sens.$ &  $Spec.$ &  $Sum$\\ 

FVT  & $(\%)$ &  $(\%)$ &  $(\%)$  &&  $(\%)$ & $(\%)$ &  $(\%)$ &&  $(\%)$ &  $(\%)$ &  $(\%)$\\ 
\midrule

$(\overrightarrow{\mu})$	&	79.91	&	79.44	&	159.34	&&	67.18	&	69.70	&	136.87	&&	71.05	&	75.14	&	146.20	\\
$(\overrightarrow{PC_{1}})$	&	75.87	&	72.37	&	148.24	&&	69.57	&	53.03	&	122.60	&&	71.76	&	65.95	&	137.70	\\
$(\overrightarrow{\mu} ,\overrightarrow{PC_{1}})$	&	81.63	&	81.42	&	\hlc[pink]{163.05}	&&	70.47	&	68.83	&	139.30	&&	75.28	&	67.28	&	142.56	\\
$(\overrightarrow{c_{1}},\overrightarrow{c_{2}})$	&	77.63	&	76.70	&	154.33	&&	70.07	&	54.15	&	124.22	&&	49.94	&	74.57	&	124.51	\\
$(\overrightarrow{c_{1}},\overrightarrow{c_{2}},\overrightarrow{PC_{1}})$	&	81.12	&	80.66	&	161.78	&&	67.53	&	63.28	&	130.80	&&	71.80	&	72.04	&	143.84	\\

\bottomrule
\end{tabular}
\label{table:bioRes2}
\end{table*}

For LDA, one can see that the vector $(\overrightarrow{\mu} ,\overrightarrow{PC_{1}})$ has the highest sum of sensitivity and specificity for both H/L and EH/EL classification, with 146.98\% and 163.05\% respectively. The vector $(\overrightarrow{\mu}) $ achieves the best performance when using kNN, where the sums of sensitivity and specificity are 130.36\% and 142.94\% for H/L and EH/EL classes respectively.

For the category Biology, binary classification results show the classifier giving the highest classification performance was LDA with the text represented by the vector $(\overrightarrow{\mu} ,\overrightarrow{PC_{1}})$ in the Original Space. This result was achieved for extreme citation classification.

\begin{table*}\centering
%\ra{1.3}
\caption{The best results of the citation classifier kNN for each FVT for H/L classification for the category Biology}
\scriptsize
\begin{tabular}{@{}lrrrrcrrrrcrrrr@{}}\toprule
 & \multicolumn{4}{c}{Original Space} & \phantom{a}& \multicolumn{4}{c}{13-Dimensional } &
\phantom{a} & \multicolumn{4}{c}{Supervised PCA}\\

 & \multicolumn{4}{c}{} & \phantom{a}& \multicolumn{4}{c}{Reduced Basis} &
\phantom{a} & \multicolumn{4}{c}{}\\
\cmidrule{2-5} \cmidrule{7-10} \cmidrule{12-15}

&$k$& $Sens.$ &  $Spec.$ &  $Sum$  &&  $k$& $Sens.$ & $Spec.$ &  $Sum$ && $k$&  $Sens.$ &  $Spec.$ &  $Sum$\\ 

FVT && $(\%)$ &  $(\%)$ &  $(\%)$  && & $(\%)$ & $(\%)$ &  $(\%)$ && & $(\%)$ &  $(\%)$ &  $(\%)$\\ 
\midrule

$(\overrightarrow{\mu})$	&	17	&	70.64	&	59.72	&	\hlc[pink]{130.36}	&&	17	&	64.39	&	61.02	&	125.40	&&	13	&	60.88	&	67.04	&	127.92	\\
$(\overrightarrow{PC_{1}})$	&	17	&	63.37	&	59.59	&	122.96	&&	15	&	28.84	&	82.18	&	111.02	&&	17	&	58.92	&	60.65	&	119.57	\\
$(\overrightarrow{\mu} ,\overrightarrow{PC_{1}})$	&	17	&	63.34	&	59.59	&	122.93	&&	17	&	64.91	&	61.99	&	126.90	&&	17	&	62.03	&	60.01	&	122.04	\\
$(\overrightarrow{c_{1}},\overrightarrow{c_{2}})$	&	11	&	65.60	&	58.36	&	123.96	&&	17	&	62.19	&	59.49	&	121.68	&&	17	&	61.11	&	58.95	&	120.06	\\
$(\overrightarrow{c_{1}},\overrightarrow{c_{2}},\overrightarrow{PC_{1}})$	&	17	&	63.99	&	59.56	&	123.55	&&	17	&	61.73	&	59.46	&	121.19	&&	17	&	60.72	&	60.71	&	121.43	\\

\bottomrule
\end{tabular}
\label{table:bioRes3}
\end{table*}

\begin{table*}\centering
%\ra{1.3}
\caption{The best results of the citation classifier kNN for each FVT for EH/EL classification for the category Biology}
\scriptsize
\begin{tabular}{@{}lrrrrcrrrrcrrrr@{}}\toprule
 & \multicolumn{4}{c}{Original Space} & \phantom{a}& \multicolumn{4}{c}{13-Dimensional } &
\phantom{a} & \multicolumn{4}{c}{Supervised PCA}\\

 & \multicolumn{4}{c}{} & \phantom{a}& \multicolumn{4}{c}{Reduced Basis} &
\phantom{a} & \multicolumn{4}{c}{}\\
\cmidrule{2-5} \cmidrule{7-10} \cmidrule{12-15}

&$k$& $Sens.$ &  $Spec.$ &  $Sum$  &&  $k$& $Sens.$ & $Spec.$ &  $Sum$ && $k$&  $Sens.$ &  $Spec.$ &  $Sum$\\ 

FVT  && $(\%)$ &  $(\%)$ &  $(\%)$  && & $(\%)$ & $(\%)$ &  $(\%)$ && & $(\%)$ &  $(\%)$ &  $(\%)$\\ 
\midrule

$(\overrightarrow{\mu})$	&	17	&	82.33	&	60.61	&	\hlc[pink]{142.94}	&&	17	&	73.21	&	64.83	&	138.03	&&	17	&	74.19	&	66.31	&	140.49	\\
$(\overrightarrow{PC_{1}})$	&	17	&	72.42	&	61.00	&	133.43	&&	17	&	66.51	&	61.69	&	128.20	&&	17	&	66.98	&	64.54	&	131.52	\\
$(\overrightarrow{\mu} ,\overrightarrow{PC_{1}})$	&	17	&	72.42	&	61.00	&	133.43	&&	17	&	72.78	&	64.25	&	137.03	&&	17	&	69.72	&	61.18	&	130.91	\\
$(\overrightarrow{c_{1}},\overrightarrow{c_{2}})$	&	17	&	71.84	&	59.63	&	131.47	&&	17	&	67.06	&	63.56	&	130.62	&&	17	&	66.94	&	62.55	&	129.49	\\
$(\overrightarrow{c_{1}},\overrightarrow{c_{2}},\overrightarrow{PC_{1}})$	&	13	&	72.39	&	61.94	&	134.33	&&	17	&	66.98	&	63.74	&	130.72	&&	17	&	70.07	&	62.16	&	132.23	\\

\bottomrule
\end{tabular}
\label{table:bioRes4}
\end{table*}

\subsubsection{\textbf{Management}\nopunct}\hspace*{\fill} \\\\
Finally, we select the Management category from the Social Science branch. We have tested two classifiers for each FVT in three spaces. The behaviours of classifiers are very similar to those in Biology. In Tables \ref{table:manRes1}-\ref{table:manRes4}, we present a comparison between LDA and kNN based on the maximum sum of sensitivity and specificity for each FVT representation and space combination separately. Overall, results similar trends to those in the category Biology. In general, we observe a significant increase in sensitivity and specificity for extreme citation classification tasks for both LDA and kNN.

According to the results in Tables \ref{table:manRes1}-\ref{table:manRes2}, LDA reaches the best classification accuracies in the Original Space. By combining the vector representations of $(\overrightarrow{\mu})$ and $(\overrightarrow{PC_{1}})$, we were able to achieve results for LDA which outperform LDA for any other FVT. The best classification accuracy of LDA is 83.22\% sensitivity and 81.81\% specificity in the binary classification of EH/EL for the vector $(\overrightarrow{\mu} ,\overrightarrow{PC_{1}})$. This is also the best accuracy achieved among the all experiments in the categories we considered. However, the difference in performance for vectors $(\overrightarrow{c_{1}},\overrightarrow{c_{2}},\overrightarrow{PC_{1}})$  is not very large, which also supports our finding that combined vectors result in an improvement of performance in the binary classification of citation.

\begin{table*}\centering

%\ra{1.3}
\caption{Results of the LDA citation classifier for H/L classification for the category Management}
\scriptsize
\begin{tabular}{@{}lrrrcrrrcrrr@{}}\toprule
 & \multicolumn{3}{c}{Original Space} & \phantom{ab}& \multicolumn{3}{c}{13-Dimensional } &
\phantom{ab} & \multicolumn{3}{c}{Supervised PCA}\\

 & \multicolumn{3}{c}{} & \phantom{ab}& \multicolumn{3}{c}{Reduced Basis} &
\phantom{ab} & \multicolumn{3}{c}{}\\
\cmidrule{2-4} \cmidrule{6-8} \cmidrule{10-12}

& $Sens.$ &  $Spec.$ &  $Sum$  &&  $Sens.$ & $Spec.$ &  $Sum$ &&  $Sens.$ &  $Spec.$ &  $Sum$\\ 

FVT  & $(\%)$ &  $(\%)$ &  $(\%)$  &&  $(\%)$ & $(\%)$ &  $(\%)$ &&  $(\%)$ &  $(\%)$ &  $(\%)$\\ 
\midrule

$(\overrightarrow{\mu})$	&	79.55	&	63.93	&	143.48	&&	70.56	&	51.25	&	121.81	&&	76.10	&	58.19	&	134.28	\\
$(\overrightarrow{PC_{1}})$	&	72.82	&	65.94	&	138.76	&&	65.45	&	49.19	&	114.64	&&	71.31	&	56.85	&	128.16	\\
$(\overrightarrow{\mu} ,\overrightarrow{PC_{1}})$	&	80.05	&	65.82	&	\hlc[pink]{145.87}	&&	69.20	&	53.18	&	122.39	&&	70.20	&	62.49	&	132.69	\\
$(\overrightarrow{c_{1}},\overrightarrow{c_{2}})$	&	79.55	&	60.96	&	140.51	&&	48.67	&	69.12	&	117.79	&&	53.23	&	62.47	&	115.70	\\
$(\overrightarrow{c_{1}},\overrightarrow{c_{2}},\overrightarrow{PC_{1}})$	&	79.89	&	65.85	&	145.74	&&	56.87	&	61.36	&	118.23	&&	78.38	&	54.54	&	132.92	\\

\bottomrule
\end{tabular}
\label{table:manRes1}
\end{table*}

\begin{table*}\centering
%\ra{1.3}
\caption{Results of the LDA citation classifier for EH/EL classification for the category Management}
\scriptsize
\begin{tabular}{@{}lrrrcrrrcrrr@{}}\toprule
 & \multicolumn{3}{c}{Original Space} & \phantom{ab}& \multicolumn{3}{c}{13-Dimensional } &
\phantom{ab} & \multicolumn{3}{c}{Supervised PCA}\\

 & \multicolumn{3}{c}{} & \phantom{ab}& \multicolumn{3}{c}{Reduced Basis} &
\phantom{ab} & \multicolumn{3}{c}{}\\
\cmidrule{2-4} \cmidrule{6-8} \cmidrule{10-12}

& $Sens.$ &  $Spec.$ &  $Sum$  &&  $Sens.$ & $Spec.$ &  $Sum$ &&  $Sens.$ &  $Spec.$ &  $Sum$\\ 

FVT  & $(\%)$ &  $(\%)$ &  $(\%)$  &&  $(\%)$ & $(\%)$ &  $(\%)$ &&  $(\%)$ &  $(\%)$ &  $(\%)$\\ 
\midrule

$(\overrightarrow{\mu})$	&	85.20	&	77.46	&	162.66	&&	72.24	&	60.17	&	132.40	&&	80.30	&	68.86	&	149.16	\\
$(\overrightarrow{PC_{1}})$	&	80.68	&	74.21	&	154.88	&&	66.86	&	54.21	&	121.07	&&	70.38	&	70.00	&	140.38	\\
$(\overrightarrow{\mu} ,\overrightarrow{PC_{1}})$	&	83.22	&	81.81	&	\hlc[pink]{165.03}	&&	67.24	&	66.56	&	133.80	&&	78.97	&	69.13	&	148.10	\\
$(\overrightarrow{c_{1}},\overrightarrow{c_{2}})$	&	81.96	&	75.82	&	157.78	&&	51.88	&	72.00	&	123.88	&&	64.27	&	61.50	&	125.77	\\
$(\overrightarrow{c_{1}},\overrightarrow{c_{2}},\overrightarrow{PC_{1}})$	&	82.21	&	82.54	&	164.75	&&	74.62	&	52.92	&	127.54	&&	75.95	&	73.77	&	149.72	\\

\bottomrule
\end{tabular}
\label{table:manRes2}
\end{table*}

Tables \ref{table:manRes3} and \ref{table:manRes4} show  the performance of kNN on binary classification tasks for H/L and EH/EL. The results in these tables indicate that kNN is also sightly better in the Original Space than in other spaces. The EL/HL classification task is performed more successfully than H/L classification by kNN. kNN performed best for EH/EL classification when using the vector $(\overrightarrow{\mu})$ in the Original Space, achieving sensitivity 82.01\% and specificity 60.09\% with sum 142.10\%.

\begin{table*}\centering
%\ra{1.3}
\caption{The best results of the citation classifier kNN for each vector according to H/L for the category Management}
\scriptsize
\begin{tabular}{@{}lrrrrcrrrrcrrrr@{}}\toprule
 & \multicolumn{4}{c}{Original Space} & \phantom{a}& \multicolumn{4}{c}{13-Dimensional } &
\phantom{a} & \multicolumn{4}{c}{Supervised PCA}\\

 & \multicolumn{4}{c}{} & \phantom{a}& \multicolumn{4}{c}{Reduced Basis} &
\phantom{a} & \multicolumn{4}{c}{}\\
\cmidrule{2-5} \cmidrule{7-10} \cmidrule{12-15}

&$k$& $Sens.$ &  $Spec.$ &  $Sum$  &&  $k$& $Sens.$ & $Spec.$ &  $Sum$ && $k$&  $Sens.$ &  $Spec.$ &  $Sum$\\ 

FVT  && $(\%)$ &  $(\%)$ &  $(\%)$  && & $(\%)$ & $(\%)$ &  $(\%)$ && & $(\%)$ &  $(\%)$ &  $(\%)$\\ 
\midrule

$(\overrightarrow{\mu})$	&	17	&	74.40	&	52.10	&	\hlc[pink]{126.50}	&&	17	&	65.09	&	52.89	&	117.98	&&	17	&	67.67	&	57.80	&	125.47	\\
$(\overrightarrow{PC_{1}})$	&	17	&	71.83	&	49.12	&	120.94	&&	17	&	61.59	&	55.39	&	116.98	&&	17	&	62.34	&	57.62	&	119.96	\\
$(\overrightarrow{\mu} ,\overrightarrow{PC_{1}})$	&	17	&	71.83	&	49.12	&	120.94	&&	17	&	68.28	&	52.80	&	121.08	&&	17	&	66.29	&	54.72	&	121.01	\\
$(\overrightarrow{c_{1}},\overrightarrow{c_{2}})$	&	13	&	65.50	&	56.01	&	121.51	&&	17	&	63.47	&	52.85	&	116.32	&&	17	&	60.71	&	56.11	&	116.82	\\
$(\overrightarrow{c_{1}},\overrightarrow{c_{2}},\overrightarrow{PC_{1}})$	&	17	&	71.74	&	49.08	&	120.81	&&	17	&	65.07	&	51.94	&	117.01	&&	17	&	65.14	&	54.98	&	120.12	\\

\bottomrule
\end{tabular}
\label{table:manRes3}
\end{table*}

\begin{table*}\centering
%\ra{1.3}
\caption{The best results for the kNN citation classifier for each FVT for EH/EL classification for the category Management}
\scriptsize
\begin{tabular}{@{}lrrrrcrrrrcrrrr@{}}\toprule
 & \multicolumn{4}{c}{Original Space} & \phantom{a}& \multicolumn{4}{c}{13-Dimensional } &
\phantom{a} & \multicolumn{4}{c}{Supervised PCA}\\

 & \multicolumn{4}{c}{} & \phantom{a}& \multicolumn{4}{c}{Reduced Basis} &
\phantom{a} & \multicolumn{4}{c}{}\\
\cmidrule{2-5} \cmidrule{7-10} \cmidrule{12-15}

&$k$& $Sens.$ &  $Spec.$ &  $Sum$  &&  $k$& $Sens.$ & $Spec.$ &  $Sum$ && $k$ &  $Sens.$ &  $Spec.$ &  $Sum$\\ 

FVT  && $(\%)$ &  $(\%)$ &  $(\%)$  && & $(\%)$ & $(\%)$ &  $(\%)$ && & $(\%)$ &  $(\%)$ &  $(\%)$\\ 
\midrule

$(\overrightarrow{\mu})$	&	17	&	82.01	&	60.09	&	\hlc[pink]{142.10}	&&	17	&	74.62	&	57.49	&	132.11	&&	17	&	78.19	&	63.31	&	141.50	\\
$(\overrightarrow{PC_{1}})$	&	17	&	80.10	&	52.92	&	133.02	&&	17	&	71.26	&	57.43	&	128.68	&&	17	&	71.93	&	60.95	&	132.88	\\
$(\overrightarrow{\mu} ,\overrightarrow{PC_{1}})$	&	17	&	80.10	&	52.94	&	133.04	&&	17	&	76.23	&	57.07	&	133.30	&&	17	&	76.26	&	57.39	&	133.65	\\
$(\overrightarrow{c_{1}},\overrightarrow{c_{2}})$	&	7	&	69.62	&	63.72	&	133.35	&&	17	&	73.39	&	56.25	&	129.64	&&	11	&	72.89	&	55.79	&	128.68	\\
$(\overrightarrow{c_{1}},\overrightarrow{c_{2}},\overrightarrow{PC_{1}})$	&	7	&	69.55	&	62.74	&	132.28	&&	17	&	74.25	&	55.26	&	129.51	&&	11	&	65.55	&	68.02	&	133.58	\\

\bottomrule
\end{tabular}
\label{table:manRes4}
\end{table*}

\begin{figure}[htb]
  \centering
  \includegraphics[height=.3\textheight, width=1\textwidth]{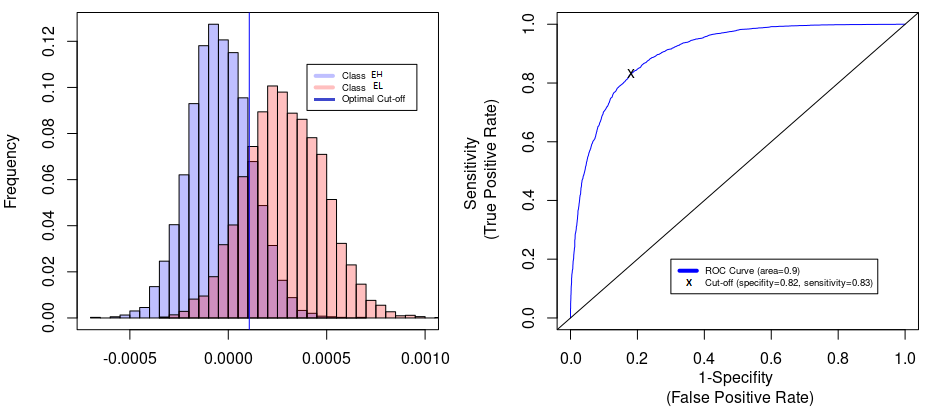}
  \caption{(Left) Data projection on the first LDA axis (Right) ROC curve for LDA. Both figures correspond to the best performance of classifiers: LDA for classes extremely little-cited (EL) and extremely highly-cited (EH) with the FVT $(\mu ,PC_{1})$ in the original space for the Management category.}
  \label{fig:BestHistogramROC}
\end{figure}

In order to illustrate the results of the best performance of classifiers, the histogram of data projection on the first LDA axis and ROC curve (see Figures \ref{fig:BestHistogramROC} are used. We note that the best performance corresponds to LDA for classes extremely little-cited (EL) and extremely highly-cited (EH) with the vector $(\mu ,PC_{1})$ in the original space of the category Management. From the histogram, we can see the separation of papers in two classes (in proportion) by the optimal cut-off found by maximising the sum of sensitivity and specificity. In the ROC curve, the corresponding AUC achieved in the classification is 90\%.

For the FVT that gives the best classification results, we applied LDA with LOOCV, with results presented in Table \ref{table:bestTable}. From the table, LDA predicted extremely  highly-cited papers with 80.43\% sensitivity and 78.90\% specificity for extremely little-cited papers with a slight difference (2.79\% for sensitivity and 2.91\% for specificity) for LDA without LOOCV. 

\begin{table*}\centering
\caption{LOOCV results for the best classification performance among all categories: LDA with LOOCV for EH/EL classification for the Management category. The results correspond to the FVT $(\mu ,PC_{1})$ in the Original Space.}
\scriptsize
\begin{tabular}{@{}lrrr@{}}\toprule
 & $Sens.$ &  $Spec.$ &  $Sum$  \\ 
FVT  & $(\%)$ &  $(\%)$ &  $(\%)$  \\ 
\midrule
$(\overrightarrow{\mu} ,\overrightarrow{PC_{1}})$	&	80.43	&	78.90	&	159.33	\\

\bottomrule
\end{tabular}
\label{table:bestTable}
\end{table*}

\subsubsection{\textbf{Summary of findings from binary classification experiments}\nopunct}\hspace*{\fill} \\\\
From the experiments we performed in the three categories, the following overall results are obtained:

\begin{enumerate}
\item In general, the highest sensitivity and specificity are reached by using the vectors $(\overrightarrow{c_{1}},\overrightarrow{c_{2}},\overrightarrow{PC_{1}})$ (for Applied Mathematics) and $(\overrightarrow{\mu} ,\overrightarrow{PC_{1}})$	(for Biology and Management) for LDA; and the vector $(\overrightarrow{\mu})$ (for all categories experimented) for kNN in almost all spaces. 

\item We observed that large $k$ values lead to relatively good performance in all classification experiments. 

\item LDA outperforms kNN in both classification tasks H/L and EH/EL in all spaces and for all FVTs. 
\item Both classifiers perform better in the EH/EL classification in all spaces and with all FVTs.

\item The best results are achieved when we employ the LDA classifier for EH/EL classification. This implies that LDA with a selection of the threshold by sensitivity+specificity maximisation is better for classifying extreme citations. LDA reached sensitivity and specificity greater than 80\%. This is a high accuracy for this type of problem. 

\item When comparing spaces, the best results are obtained for the Original Space in almost all experiments, followed by the space constructed by Supervised PCA, and then the space constructed with 13-Dimensional reduced basis. We can conclude that representing words in 13-dimensional space leads to some loss of information about the citation of papers.
\item Semantics is a more important indicator for citation count for some categories than others. For the Biology and Management categories  the results show an improvement over those for Applied Mathematics. The poorest results were obtained for the category Applied Mathematics.
\item Finally, combining $(\overrightarrow{PC_{1}})$ vector with vectors $(\overrightarrow{\mu})$ and $(\overrightarrow{c_{1}},\overrightarrow{c_{2}})$  results in an improvement of the classifier performance.  
  
\end{enumerate}

\section{Conclusion and Discussion} \label{concDiss}
In this paper, two issues in Natural Language Processing (NLP) have been identified, solutions have been suggested and  proposed approaches have been analysed. Firstly, we introduced a new text representation method which is one of the main problems for all NLP applications. Secondly, we evaluated the impact of scientific articles using this new representation technique. Predictive models are studied for classifying the citation of papers based on their informational semantics and several representation models in different vector spaces are tested in classification experiments. The aim of this study is to assess the discrimination ability of scientific success of papers through their semantics.
  
A new informational text representation model is introduced. The basis of the model is the corpus LSC and words which are represented in the Meaning Space (MS) according to their importance as extracted from their usage in subject categories \cite{nesli2,nesli}. In this approach, each text is a cloud of words represented by vectors of Relative Information Gains (RIGs) from word to category in the MS. In order to construct a text representation, we have used the distribution of the corresponding words' RIGs in each dimension. The Feature Vectors of Text (FVTs) are created using sets of parameters (e.g. mean values of a cloud's points) summarising the information in texts. Five different vector representations  are explored, using combinations of the mean vector, the vector of the first principal component for each text and two centroid vectors obtained by the $k$-means clustering of words in the text, are introduced as informational representations of semantics and analysed for the selected binary classification problem. 

With the FVTs constructed, we have showed how much information the semantics in texts carry for the citation of scientific articles. Classification is performed for a binary problem: classification of highly-cited papers and little-cited papers (classes H/L). We also conduct experiments to classify papers for two more extreme citation counts: extremely highly-cited (EH) and extremely little-cited (EL) papers. Distinguishing between two classes was done by relative thresholds defined for each scientific category individually. Two classification methods were investigated for comparison: LDA and kNN. Five different vector representations were tested with binary classifiers. All experiments were also repeated in three spaces constructed: the Original Space, 13-Dimensional reduced basis and the space constructed after supervised PCA. Our experiments use three categories selected from three branches of science: Applied Mathematics, Biology and Management. The study underlines the importance of vector representations and the spaces to represent texts in prediction citation, and comparison of categories having different characteristics for citation.

We found that informational semantics is a very promising approach for developing a quantitative evaluation and predictive model of citation count. The results of binary classification show that the best performance is achieved for extreme citation prediction by LDA in the Original Space: identification of extremely highly cited papers with 83.22\% sensitivity and identification of extremely less-cited papers with 81.81\% specificity. LDA outperforms kNN for all categories, vectors and spaces tested.

The prediction of citation in Applied Mathematics is inherently more difficult than for Biology and Management since the category is heterogeneous, meaning that it contains papers from different disciplines and citation patterns are different for these individual fields. Thus Applied Mathematics is not well-separated into popular scientific fields and so contains a mixture of category-based semantics. The categories of Biology and Management are more homogeneous in terms of the papers assigned to them, and  have a clear separation into top popular areas. Therefore, these two categories are expected to reach higher classification scores than Applied Mathematics. 

A very important finding from the binary classification experiments is that extreme classes are more discriminative than other, achieving higher sensitivity and specificity in all experiments. This can be explained by the fact that two extremes indicate the agreement among scientific community as `containing information useful for many researchers' and `being outside of the interest of the community' and the semantics in papers is an important indicator for scientific excellence of a paper. 

As a result, informational text representations proved to be efficient in the cases where binary classification was performed in distinguishing the impact of articles. We have shown that clues -- extracted from the importance of words in categories -- about the context of a paper provide important information about citation differentiation in scientific articles. This fact is much more clear for some scientific categories than others. 

Multi-class classification of citation through informational semantics has not yet been assessed. We recommend further and more in-depth research to compare multi-class classification tasks with a combination of different vector representations. Beside the category-based predictors, we also encourage further research to compare mixture of categories.

\section*{Appendices}
\appendix
\section{List of categories with the descriptive statistics obtained for the number of citation for each of categories. Categories are sorted by the average citation count assigned to the category }\label{DesStatsCats}
\raggedbottom	
\scriptsize
\begin{longtable}[h]{L{4.2cm} R{1.2cm} R{.7cm} R{.6cm} R{.7cm}  R{.4cm}  R{.4cm} R{.4cm}  R{.7cm} R{.7cm}}
	
		\caption{List of categories with the descriptive statistics obtained for the number of citation}\\ 
		\hline
  	  
\multicolumn{1}{C{4.2cm}}{\textbf{Set}} & \multicolumn{1}{C{1.2cm}}{\textbf{$N$}} & \multicolumn{1}{C{.7cm}}{\textbf{Max}} &\multicolumn{1}{C{.6cm}}{\textbf{Min}}  & \multicolumn{1}{C{.7cm}}{\textbf{$\mu$}}  & \multicolumn{1}{C{.4cm}}{\textbf{$Q_{1}$}} &\multicolumn{1}{C{.4cm}}{\textbf{$Q_{2}$}} & \multicolumn{1}{C{.4cm}}{\textbf{$Q_{3}$}} & \multicolumn{1}{C{,7cm}}{\textbf{$\sigma$}} &\multicolumn{1}{C{.7cm}}{\textbf{SE}} \\ \hline

\endhead
\hline
\endfoot

Cell Biology	&	 23,108 	&	 978 	&	0	&	20.67	&	6	&	12	&	24	&	32.37	&	0.213	\\
Chemistry, Multidisciplinary	&	 55,907 	&	 2,210 	&	0	&	18.90	&	3	&	9	&	21	&	37.29	&	0.158	\\
Multidisciplinary Sciences	&	 53,140 	&	 3,004 	&	0	&	18.32	&	4	&	9	&	18	&	44.42	&	0.193	\\
Chemistry, Physical	&	 58,065 	&	 2,288 	&	0	&	18.11	&	5	&	10	&	20	&	32.86	&	0.136	\\
Nanoscience \& Nanotechnology	&	 35,050 	&	 2,464 	&	0	&	18.11	&	2	&	8	&	20	&	40.29	&	0.215	\\
Critical Care Medicine	&	 3,982 	&	 320 	&	0	&	17.53	&	5	&	11	&	21	&	22.63	&	0.359	\\
Allergy	&	 1,765 	&	 488 	&	0	&	17.41	&	4	&	10	&	20	&	26.99	&	0.642	\\
Neuroimaging	&	 2,702 	&	 527 	&	0	&	17.24	&	6	&	12	&	22	&	23.59	&	0.454	\\
Cell \& Tissue Engineering	&	 2,455 	&	 261 	&	0	&	16.81	&	4	&	10	&	20	&	24.64	&	0.497	\\
Medicine, General \& Internal	&	 16,179 	&	 3,908 	&	0	&	16.01	&	2	&	4	&	10	&	68.91	&	0.542	\\
Gastroenterology \& Hepatology	&	 10,943 	&	 667 	&	0	&	15.50	&	4	&	9	&	18	&	24.99	&	0.239	\\
Oncology	&	 34,339 	&	 8,234 	&	0	&	15.24	&	4	&	9	&	17	&	52.34	&	0.282	\\
Hematology	&	 9,096 	&	 399 	&	0	&	15.17	&	4	&	9	&	18	&	22.19	&	0.233	\\
Endocrinology \& Metabolism	&	 14,622 	&	 1,388 	&	0	&	14.95	&	5	&	10	&	18	&	22.18	&	0.183	\\
Genetics \& Heredity	&	 19,512 	&	 6,756 	&	0	&	14.94	&	4	&	8	&	17	&	59.04	&	0.423	\\
Materials Science, Biomaterials	&	 8,040 	&	 280 	&	0	&	14.74	&	4	&	10	&	19	&	17.46	&	0.195	\\
Electrochemistry	&	 15,663 	&	 290 	&	0	&	14.56	&	4	&	10	&	19	&	16.46	&	0.132	\\
Biochemistry \& Molecular Biology	&	 47,490 	&	 1,829 	&	0	&	14.56	&	4	&	9	&	16	&	30.65	&	0.141	\\
Immunology	&	 18,270 	&	 515 	&	0	&	14.52	&	4	&	9	&	17	&	21.29	&	0.157	\\
Peripheral Vascular Disease	&	 8,700 	&	 1,104 	&	0	&	14.19	&	4	&	9	&	16	&	24.90	&	0.267	\\
Rheumatology	&	 3,942 	&	 481 	&	0	&	14.16	&	4	&	9	&	17	&	19.92	&	0.317	\\
Neurosciences	&	 32,972 	&	 808 	&	0	&	14.15	&	5	&	9	&	17	&	18.72	&	0.103	\\
Green \& Sustainable Science \& Technology	&	 6,412 	&	 304 	&	0	&	14.10	&	3	&	9	&	19	&	17.12	&	0.214	\\
Respiratory System	&	 7,666 	&	 800 	&	0	&	14.08	&	4	&	9	&	17	&	21.08	&	0.241	\\
Virology	&	 6,270 	&	 680 	&	0	&	14.06	&	5	&	9	&	17	&	17.98	&	0.227	\\
Cardiac \& Cardiovascular Systems	&	 16,369 	&	 813 	&	0	&	14.02	&	3	&	8	&	16	&	22.63	&	0.177	\\
Evolutionary Biology	&	 5,742 	&	 1,085 	&	0	&	13.96	&	5	&	9	&	17	&	21.70	&	0.286	\\
Engineering, Environmental	&	 14,614 	&	 367 	&	0	&	13.87	&	2	&	8	&	19	&	18.81	&	0.156	\\
Physics, Condensed Matter	&	 27,316 	&	 2,288 	&	0	&	13.83	&	2	&	6	&	14	&	35.28	&	0.213	\\
Microbiology	&	 17,252 	&	 680 	&	0	&	13.73	&	4	&	9	&	17	&	18.98	&	0.145	\\
Biochemical Research Methods	&	 15,050 	&	 4,744 	&	0	&	13.56	&	3	&	8	&	15	&	65.25	&	0.532	\\
Agricultural Engineering	&	 3,727 	&	 496 	&	0	&	13.50	&	3	&	9	&	19	&	16.76	&	0.274	\\
Astronomy \& Astrophysics	&	 22,825 	&	 1,243 	&	0	&	13.24	&	2	&	7	&	16	&	24.20	&	0.160	\\
Nutrition \& Dietetics	&	 9,416 	&	 244 	&	0	&	13.11	&	4	&	9	&	17	&	14.13	&	0.146	\\
Engineering, Chemical	&	 29,171 	&	 1,105 	&	0	&	13.06	&	3	&	7	&	16	&	22.10	&	0.129	\\
Developmental Biology	&	 3,593 	&	 217 	&	0	&	13.04	&	4	&	8	&	16	&	15.59	&	0.260	\\
Infectious Diseases	&	 12,521 	&	 515 	&	0	&	12.74	&	4	&	8	&	15	&	18.28	&	0.163	\\
Biotechnology \& Applied Microbiology	&	 26,286 	&	 4,744 	&	0	&	12.70	&	3	&	7	&	14	&	49.30	&	0.304	\\
Ecology	&	 16,760 	&	 392 	&	0	&	12.55	&	4	&	8	&	16	&	15.76	&	0.122	\\
Meteorology \& Atmospheric Sciences	&	 12,318 	&	 1,659 	&	0	&	12.42	&	3	&	8	&	15	&	24.22	&	0.218	\\
Clinical Neurology	&	 22,127 	&	 1,104 	&	0	&	12.32	&	4	&	8	&	15	&	18.68	&	0.126	\\
Environmental Sciences	&	 42,082 	&	 1,105 	&	0	&	12.28	&	3	&	7	&	15	&	21.18	&	0.103	\\
Chemistry, Analytical	&	 21,490 	&	 290 	&	0	&	12.14	&	4	&	8	&	15	&	14.68	&	0.100	\\
Geriatrics \& Gerontology	&	 4,742 	&	 513 	&	0	&	12.13	&	4	&	8	&	15	&	15.78	&	0.229	\\
Psychiatry	&	 16,055 	&	 317 	&	0	&	12.11	&	3	&	8	&	15	&	15.51	&	0.122	\\
Psychology, Developmental	&	 4,390 	&	 306 	&	0	&	11.83	&	4	&	8	&	15	&	13.60	&	0.205	\\
Anesthesiology	&	 2,943 	&	 326 	&	0	&	11.82	&	4	&	8	&	15	&	14.57	&	0.269	\\
Energy \& Fuels	&	 44,202 	&	 1,105 	&	0	&	11.81	&	0	&	5	&	16	&	22.14	&	0.105	\\
Parasitology	&	 5,683 	&	 680 	&	0	&	11.79	&	4	&	8	&	14	&	17.54	&	0.233	\\
Urology \& Nephrology	&	 8,264 	&	 994 	&	0	&	11.75	&	3	&	7	&	14	&	19.56	&	0.215	\\
Physics, Particles \& Fields	&	 13,203 	&	 1,508 	&	0	&	11.72	&	2	&	6	&	14	&	22.81	&	0.199	\\
Chemistry, Organic	&	 17,941 	&	 210 	&	0	&	11.61	&	3	&	8	&	15	&	13.11	&	0.098	\\
Chemistry, Applied	&	 14,058 	&	 174 	&	0	&	11.50	&	3	&	8	&	15	&	12.27	&	0.104	\\
Physics, Atomic, Molecular \& Chemical	&	 17,010 	&	 1,143 	&	0	&	11.46	&	3	&	7	&	14	&	21.09	&	0.162	\\
Environmental Studies	&	 7,811 	&	 775 	&	0	&	11.30	&	2	&	6	&	13	&	21.16	&	0.239	\\
Medicine, Research \& Experimental	&	 19,744 	&	 978 	&	0	&	11.24	&	2	&	6	&	13	&	21.63	&	0.154	\\
Toxicology	&	 9,613 	&	 338 	&	0	&	11.17	&	4	&	8	&	14	&	12.80	&	0.131	\\
Polymer Science	&	 18,017 	&	 358 	&	0	&	11.15	&	3	&	7	&	15	&	14.02	&	0.104	\\
Materials Science, Multidisciplinary	&	 112,912 	&	 2,464 	&	0	&	10.99	&	0	&	4	&	12	&	27.89	&	0.083	\\
Biophysics	&	 12,630 	&	 488 	&	0	&	10.97	&	3	&	7	&	14	&	13.87	&	0.123	\\
Geochemistry \& Geophysics	&	 10,023 	&	 213 	&	0	&	10.92	&	3	&	7	&	14	&	13.30	&	0.133	\\
Psychology, Experimental	&	 6,784 	&	 572 	&	0	&	10.75	&	3	&	7	&	14	&	14.55	&	0.177	\\
Psychology, Clinical	&	 6,860 	&	 282 	&	0	&	10.57	&	3	&	7	&	14	&	13.03	&	0.157	\\
Psychology	&	 6,989 	&	 300 	&	0	&	10.56	&	3	&	7	&	14	&	14.75	&	0.176	\\
Transplantation	&	 4,105 	&	 559 	&	0	&	10.47	&	3	&	6	&	13	&	17.34	&	0.271	\\
Physiology	&	 9,009 	&	 231 	&	0	&	10.47	&	4	&	8	&	13	&	11.12	&	0.117	\\
Pharmacology \& Pharmacy	&	 30,713 	&	 323 	&	0	&	10.47	&	3	&	7	&	14	&	11.92	&	0.068	\\
Gerontology	&	 2,531 	&	 513 	&	0	&	10.42	&	3	&	7	&	13	&	17.23	&	0.342	\\
Reproductive Biology	&	 3,984 	&	 454 	&	0	&	10.40	&	4	&	7	&	13	&	13.95	&	0.221	\\
Behavioral Sciences	&	 5,922 	&	 186 	&	0	&	10.37	&	4	&	8	&	14	&	9.92	&	0.129	\\
Biology	&	 9,917 	&	 517 	&	0	&	10.36	&	2	&	6	&	13	&	16.08	&	0.162	\\
Biodiversity Conservation	&	 4,705 	&	 352 	&	0	&	10.31	&	2	&	6	&	13	&	14.79	&	0.216	\\
Chemistry, Medicinal	&	 12,456 	&	 530 	&	0	&	10.31	&	4	&	7	&	13	&	12.84	&	0.115	\\
Substance Abuse	&	 3,433 	&	 226 	&	0	&	10.29	&	3	&	7	&	13	&	12.62	&	0.215	\\
Geography	&	 3,908 	&	 775 	&	0	&	10.27	&	2	&	6	&	13	&	18.95	&	0.303	\\
Psychology, Mathematical	&	 538 	&	 572 	&	0	&	10.25	&	2	&	5	&	10	&	31.19	&	1.345	\\
Materials Science, Composites	&	 4,277 	&	 357 	&	0	&	10.11	&	2	&	5	&	13	&	15.48	&	0.237	\\
Geography, Physical	&	 6,806 	&	 378 	&	0	&	10.09	&	2	&	6	&	13	&	16.09	&	0.195	\\
Food Science \& Technology	&	 20,414 	&	 499 	&	0	&	9.86	&	3	&	7	&	13	&	11.06	&	0.077	\\
Materials Science, Coatings \& Films	&	 7,226 	&	 129 	&	0	&	9.78	&	3	&	7	&	13	&	10.76	&	0.127	\\
Orthopedics	&	 10,538 	&	 234 	&	0	&	9.76	&	3	&	7	&	13	&	11.35	&	0.111	\\
Pathology	&	 7,217 	&	 373 	&	0	&	9.76	&	3	&	6	&	12	&	13.98	&	0.165	\\
Ophthalmology	&	 7,830 	&	 615 	&	0	&	9.75	&	3	&	6	&	12	&	15.19	&	0.172	\\
Sport Sciences	&	 8,368 	&	 184 	&	0	&	9.69	&	3	&	7	&	13	&	11.48	&	0.126	\\
Radiology, Nuclear Medicine \& Medical Imaging	&	 21,014 	&	 527 	&	0	&	9.69	&	2	&	6	&	12	&	14.58	&	0.101	\\
Psychology, Biological	&	 1,527 	&	 294 	&	0	&	9.67	&	3	&	7	&	12	&	13.22	&	0.338	\\
Mathematical \& Computational Biology	&	 8,015 	&	 4,744 	&	0	&	9.67	&	1	&	4	&	9	&	76.31	&	0.852	\\
Mycology	&	 1,829 	&	 232 	&	0	&	9.58	&	3	&	6	&	12	&	13.51	&	0.316	\\
Plant Sciences	&	 21,321 	&	 449 	&	0	&	9.54	&	2	&	6	&	12	&	12.81	&	0.088	\\
Tropical Medicine	&	 3,696 	&	 318 	&	0	&	9.36	&	3	&	6	&	12	&	12.96	&	0.213	\\
Surgery	&	 30,805 	&	 559 	&	0	&	9.35	&	2	&	6	&	12	&	12.95	&	0.074	\\
Limnology	&	 1,941 	&	 217 	&	0	&	9.35	&	3	&	6	&	12	&	12.16	&	0.276	\\
Thermodynamics	&	 13,852 	&	 238 	&	0	&	9.33	&	1	&	5	&	13	&	12.56	&	0.107	\\
Obstetrics \& Gynecology	&	 9,883 	&	 454 	&	0	&	9.32	&	3	&	6	&	12	&	12.25	&	0.123	\\
Psychology, Applied	&	 3,523 	&	 404 	&	0	&	9.27	&	2	&	6	&	12	&	15.31	&	0.258	\\
Geosciences, Multidisciplinary	&	 24,644 	&	 383 	&	0	&	9.26	&	2	&	6	&	12	&	13.90	&	0.089	\\
Mineralogy	&	 2,550 	&	 139 	&	0	&	9.24	&	3	&	6	&	12	&	10.52	&	0.208	\\
Physics, Multidisciplinary	&	 22,930 	&	 1,178 	&	0	&	9.21	&	1	&	3	&	9	&	23.95	&	0.158	\\
Health Care Sciences \& Services	&	 7,999 	&	 288 	&	0	&	9.21	&	3	&	6	&	12	&	12.78	&	0.143	\\
Chemistry, Inorganic \& Nuclear	&	 12,591 	&	 173 	&	0	&	9.19	&	2	&	6	&	12	&	11.15	&	0.099	\\
Psychology, Social	&	 3,548 	&	 193 	&	0	&	9.15	&	3	&	6	&	11	&	11.46	&	0.192	\\
Psychology, Educational	&	 2,112 	&	 151 	&	0	&	9.09	&	2	&	6	&	12	&	12.17	&	0.265	\\
Soil Science	&	 4,800 	&	 149 	&	0	&	9.04	&	2	&	5	&	12	&	11.85	&	0.171	\\
Psychology, Multidisciplinary	&	 8,332 	&	 717 	&	0	&	8.84	&	1	&	5	&	10	&	17.32	&	0.190	\\
Public, Environmental \& Occupational Health	&	 25,493 	&	 655 	&	0	&	8.77	&	2	&	5	&	11	&	13.79	&	0.086	\\
Water Resources	&	 13,997 	&	 367 	&	0	&	8.76	&	2	&	5	&	11	&	11.81	&	0.100	\\
Physics, Applied	&	 78,796 	&	 2,288 	&	0	&	8.73	&	0	&	3	&	9	&	24.73	&	0.088	\\
Dermatology	&	 5,793 	&	 249 	&	0	&	8.58	&	2	&	5	&	11	&	11.86	&	0.156	\\
Marine \& Freshwater Biology	&	 10,124 	&	 157 	&	0	&	8.53	&	3	&	6	&	11	&	9.66	&	0.096	\\
Instruments \& Instrumentation	&	 17,090 	&	 419 	&	0	&	8.48	&	1	&	4	&	10	&	14.16	&	0.108	\\
Pediatrics	&	 13,364 	&	 309 	&	0	&	8.43	&	2	&	5	&	10	&	12.32	&	0.107	\\
Health Policy \& Services	&	 5,318 	&	 159 	&	0	&	8.42	&	2	&	6	&	11	&	10.38	&	0.142	\\
Social Sciences, Biomedical	&	 3,003 	&	 132 	&	0	&	8.40	&	3	&	6	&	11	&	9.50	&	0.173	\\
Urban Studies	&	 2,309 	&	 334 	&	0	&	8.11	&	1	&	4	&	10	&	14.43	&	0.300	\\
Medical Laboratory Technology	&	 2,598 	&	 337 	&	0	&	8.05	&	2	&	5	&	10	&	11.81	&	0.232	\\
Ergonomics	&	 1,431 	&	 109 	&	0	&	7.94	&	3	&	6	&	10	&	8.76	&	0.232	\\
Metallurgy \& Metallurgical Engineering	&	 16,898 	&	 209 	&	0	&	7.92	&	1	&	4	&	10	&	11.09	&	0.085	\\
Integrative \& Complementary Medicine	&	 3,453 	&	 88 	&	0	&	7.91	&	2	&	6	&	11	&	7.80	&	0.133	\\
Geology	&	 2,153 	&	 108 	&	0	&	7.86	&	2	&	5	&	10	&	9.18	&	0.198	\\
Emergency Medicine	&	 2,627 	&	 143 	&	0	&	7.82	&	2	&	5	&	10	&	10.49	&	0.205	\\
Engineering, Biomedical	&	 17,786 	&	 368 	&	0	&	7.82	&	0	&	3	&	10	&	13.45	&	0.101	\\
Dentistry, Oral Surgery \& Medicine	&	 8,502 	&	 445 	&	0	&	7.80	&	2	&	5	&	10	&	10.12	&	0.110	\\
Spectroscopy	&	 7,388 	&	 173 	&	0	&	7.75	&	2	&	5	&	10	&	9.13	&	0.106	\\
Forestry	&	 4,472 	&	 148 	&	0	&	7.72	&	2	&	5	&	10	&	8.95	&	0.134	\\
Crystallography	&	 6,932 	&	 743 	&	0	&	7.71	&	2	&	5	&	10	&	13.28	&	0.160	\\
Materials Science, Ceramics	&	 6,222 	&	 257 	&	0	&	7.69	&	1	&	5	&	11	&	9.85	&	0.125	\\
Statistics \& Probability	&	 9,532 	&	 4,744 	&	0	&	7.65	&	1	&	3	&	7	&	68.73	&	0.704	\\
Physics, Fluids \& Plasmas	&	 9,704 	&	 229 	&	0	&	7.64	&	2	&	5	&	10	&	10.26	&	0.104	\\
Fisheries	&	 4,702 	&	 152 	&	0	&	7.61	&	2	&	5	&	10	&	9.42	&	0.137	\\
Demography	&	 948 	&	 197 	&	0	&	7.46	&	2	&	5	&	10	&	10.29	&	0.334	\\
Andrology	&	 391 	&	 36 	&	0	&	7.35	&	3	&	6	&	10	&	6.00	&	0.303	\\
Oceanography	&	 7,417 	&	 200 	&	0	&	7.33	&	1	&	5	&	10	&	9.74	&	0.113	\\
Primary Health Care	&	 1,269 	&	 253 	&	0	&	7.29	&	2	&	5	&	10	&	10.76	&	0.302	\\
Audiology \& Speech-Language Pathology	&	 2,052 	&	 95 	&	0	&	7.23	&	3	&	5	&	9	&	7.64	&	0.169	\\
Medicine, Legal	&	 1,711 	&	 139 	&	0	&	7.21	&	2	&	5	&	9	&	9.62	&	0.233	\\
Business	&	 9,394 	&	 404 	&	0	&	7.19	&	0	&	2	&	9	&	14.92	&	0.154	\\
Agriculture, Multidisciplinary	&	 6,406 	&	 203 	&	0	&	7.16	&	1	&	4	&	10	&	9.49	&	0.119	\\
Paleontology	&	 2,503 	&	 92 	&	0	&	7.09	&	2	&	5	&	9	&	7.31	&	0.146	\\
Family Studies	&	 2,229 	&	 80 	&	0	&	7.05	&	2	&	5	&	9	&	7.59	&	0.161	\\
Political Science	&	 5,106 	&	 293 	&	0	&	7.04	&	1	&	4	&	9	&	10.89	&	0.152	\\
Operations Research \& Management Science	&	 11,879 	&	 271 	&	0	&	6.99	&	0	&	3	&	9	&	12.09	&	0.111	\\
Agronomy	&	 8,651 	&	 148 	&	0	&	6.98	&	1	&	4	&	9	&	9.42	&	0.101	\\
Otorhinolaryngology	&	 4,797 	&	 432 	&	0	&	6.96	&	2	&	5	&	9	&	9.65	&	0.139	\\
Management	&	 14,339 	&	 304 	&	0	&	6.92	&	0	&	2	&	8	&	13.30	&	0.111	\\
Imaging Science \& Photographic Technology	&	 9,353 	&	 737 	&	0	&	6.91	&	0	&	2	&	7	&	17.96	&	0.186	\\
Anthropology	&	 3,149 	&	 84 	&	0	&	6.86	&	1	&	4	&	9	&	8.92	&	0.159	\\
Planning \& Development	&	 4,115 	&	 147 	&	0	&	6.81	&	0	&	3	&	9	&	11.38	&	0.177	\\
Transportation	&	 4,035 	&	 113 	&	0	&	6.80	&	1	&	4	&	9	&	8.93	&	0.141	\\
Hospitality, Leisure, Sport \& Tourism	&	 2,998 	&	 144 	&	0	&	6.80	&	0	&	4	&	9	&	10.55	&	0.193	\\
Physics, Mathematical	&	 10,426 	&	 412 	&	0	&	6.79	&	1	&	4	&	8	&	11.69	&	0.114	\\
Rehabilitation	&	 7,791 	&	 182 	&	0	&	6.78	&	2	&	5	&	9	&	7.84	&	0.089	\\
Information Science \& Library Science	&	 4,565 	&	 162 	&	0	&	6.62	&	0	&	3	&	8	&	11.35	&	0.168	\\
Agricultural Economics \& Policy	&	 880 	&	 143 	&	0	&	6.62	&	1	&	3	&	8	&	11.32	&	0.382	\\
Criminology \& Penology	&	 2,015 	&	 91 	&	0	&	6.60	&	2	&	4	&	9	&	8.41	&	0.187	\\
Physics, Nuclear	&	 7,876 	&	 324 	&	0	&	6.58	&	1	&	3	&	8	&	12.36	&	0.139	\\
Communication	&	 3,200 	&	 169 	&	0	&	6.55	&	1	&	3	&	8	&	10.80	&	0.191	\\
Sociology	&	 4,725 	&	 192 	&	0	&	6.54	&	1	&	4	&	8	&	9.17	&	0.133	\\
Medical Informatics	&	 3,991 	&	 368 	&	0	&	6.54	&	0	&	3	&	8	&	12.07	&	0.191	\\
Materials Science, Paper \& Wood	&	 1,963 	&	 358 	&	0	&	6.51	&	2	&	4	&	8	&	10.92	&	0.247	\\
Materials Science, Textiles	&	 2,548 	&	 358 	&	0	&	6.19	&	1	&	3	&	8	&	10.78	&	0.214	\\
Economics	&	 22,338 	&	 217 	&	0	&	6.11	&	1	&	3	&	7	&	10.78	&	0.072	\\
Social Issues	&	 1,296 	&	 72 	&	0	&	6.04	&	2	&	4	&	8	&	7.09	&	0.197	\\
Engineering, Geological	&	 4,573 	&	 222 	&	0	&	6.00	&	0	&	2	&	8	&	10.39	&	0.154	\\
Zoology	&	 11,218 	&	 234 	&	0	&	6.00	&	2	&	4	&	8	&	7.50	&	0.071	\\
Anatomy \& Morphology	&	 1,889 	&	 62 	&	0	&	5.97	&	2	&	4	&	8	&	6.95	&	0.160	\\
Agriculture, Dairy \& Animal Science	&	 6,163 	&	 289 	&	0	&	5.96	&	1	&	4	&	8	&	7.58	&	0.097	\\
Education, Special	&	 1,666 	&	 56 	&	0	&	5.95	&	1	&	4	&	8	&	6.71	&	0.164	\\
Remote Sensing	&	 11,388 	&	 469 	&	0	&	5.94	&	0	&	1	&	6	&	13.97	&	0.131	\\
Mathematics, Interdisciplinary Applications	&	 10,072 	&	 572 	&	0	&	5.89	&	1	&	2	&	7	&	13.36	&	0.133	\\
International Relations	&	 2,941 	&	 125 	&	0	&	5.88	&	1	&	3	&	7	&	8.70	&	0.160	\\
Entomology	&	 5,704 	&	 116 	&	0	&	5.88	&	2	&	4	&	8	&	7.38	&	0.098	\\
Engineering, Civil	&	 22,127 	&	 222 	&	0	&	5.83	&	0	&	2	&	8	&	9.93	&	0.067	\\
Social Work	&	 2,114 	&	 78 	&	0	&	5.82	&	2	&	4	&	8	&	6.69	&	0.145	\\
Nursing	&	 6,637 	&	 101 	&	0	&	5.72	&	2	&	4	&	8	&	6.42	&	0.079	\\
Computer Science, Interdisciplinary Applications	&	 29,153 	&	 4,744 	&	0	&	5.67	&	0	&	1	&	6	&	40.84	&	0.239	\\
Construction \& Building Technology	&	 12,078 	&	 167 	&	0	&	5.64	&	0	&	1	&	7	&	9.91	&	0.090	\\
Acoustics	&	 6,935 	&	 236 	&	0	&	5.60	&	0	&	2	&	7	&	10.04	&	0.121	\\
Industrial Relations \& Labor	&	 879 	&	 57 	&	0	&	5.59	&	1	&	3	&	7	&	7.48	&	0.252	\\
Women's Studies	&	 1,341 	&	 81 	&	0	&	5.50	&	1	&	3	&	7	&	7.26	&	0.198	\\
Ethics	&	 1,928 	&	 61 	&	0	&	5.46	&	1	&	3	&	7	&	7.49	&	0.170	\\
Optics	&	 47,737 	&	 1,277 	&	0	&	5.44	&	0	&	2	&	6	&	14.16	&	0.065	\\
Veterinary Sciences	&	 11,502 	&	 289 	&	0	&	5.42	&	1	&	3	&	7	&	7.17	&	0.067	\\
Archaeology	&	 2,118 	&	 89 	&	0	&	5.41	&	1	&	3	&	7	&	7.45	&	0.162	\\
Engineering, Industrial	&	 10,362 	&	 532 	&	0	&	5.36	&	0	&	1	&	6	&	11.99	&	0.118	\\
Public Administration	&	 2,204 	&	 81 	&	0	&	5.34	&	0	&	2	&	7	&	8.54	&	0.182	\\
Mechanics	&	 33,545 	&	 229 	&	0	&	5.33	&	0	&	1	&	7	&	9.77	&	0.053	\\
Computer Science, Artificial Intelligence	&	 41,210 	&	 2,100 	&	0	&	5.31	&	0	&	1	&	5	&	20.59	&	0.101	\\
Microscopy	&	 1,319 	&	 129 	&	0	&	5.16	&	1	&	3	&	7	&	7.98	&	0.220	\\
Ethnic Studies	&	 675 	&	 42 	&	0	&	5.14	&	1	&	3	&	7	&	5.76	&	0.222	\\
Medical Ethics	&	 674 	&	 57 	&	0	&	5.08	&	1	&	3	&	7	&	6.05	&	0.233	\\
Automation \& Control Systems	&	 29,427 	&	 2,100 	&	0	&	5.02	&	0	&	1	&	4	&	18.32	&	0.107	\\
Ornithology	&	 1,008 	&	 70 	&	0	&	4.92	&	1	&	4	&	7	&	5.54	&	0.175	\\
Computer Science, Cybernetics	&	 3,652 	&	 205 	&	0	&	4.88	&	0	&	1	&	4.25	&	12.22	&	0.202	\\
Mining \& Mineral Processing	&	 2,687 	&	 81 	&	0	&	4.70	&	0	&	2	&	6	&	7.39	&	0.143	\\
Telecommunications	&	 40,550 	&	 2,028 	&	0	&	4.50	&	0	&	1	&	3	&	21.47	&	0.107	\\
Engineering, Manufacturing	&	 13,102 	&	 236 	&	0	&	4.48	&	0	&	1	&	6	&	8.88	&	0.078	\\
Business, Finance	&	 7,214 	&	 162 	&	0	&	4.48	&	0	&	1	&	5	&	9.24	&	0.109	\\
Transportation Science \& Technology	&	 8,411 	&	 201 	&	0	&	4.39	&	0	&	1	&	5	&	9.97	&	0.109	\\
Mathematics, Applied	&	 27,982 	&	 253 	&	0	&	4.33	&	0	&	2	&	5	&	8.03	&	0.048	\\
Engineering, Multidisciplinary	&	 21,144 	&	 696 	&	0	&	4.26	&	0	&	1	&	4	&	10.58	&	0.073	\\
Social Sciences, Mathematical Methods	&	 3,496 	&	 225 	&	0	&	4.21	&	0	&	1	&	5	&	9.46	&	0.160	\\
Horticulture	&	 5,338 	&	 105 	&	0	&	4.19	&	0	&	2	&	5	&	6.75	&	0.092	\\
Law	&	 3,574 	&	 77 	&	0	&	4.03	&	0	&	2	&	5	&	6.55	&	0.110	\\
Computer Science, Software Engineering	&	 17,103 	&	 696 	&	0	&	4.00	&	0	&	1	&	4	&	10.52	&	0.080	\\
Nuclear Science \& Technology	&	 11,359 	&	 229 	&	0	&	4.00	&	0	&	2	&	5	&	6.72	&	0.063	\\
Linguistics	&	 5,921 	&	 112 	&	0	&	3.96	&	0	&	2	&	5	&	6.47	&	0.084	\\
Engineering, Ocean	&	 2,352 	&	 200 	&	0	&	3.83	&	0	&	0	&	5	&	8.62	&	0.178	\\
Materials Science, Characterization \& Testing	&	 3,878 	&	 246 	&	0	&	3.82	&	0	&	1	&	5	&	7.44	&	0.119	\\
Engineering, Electrical \& Electronic	&	 174,272 	&	 2,028 	&	0	&	3.80	&	0	&	0	&	3	&	13.96	&	0.033	\\
Engineering, Aerospace	&	 4,435 	&	 270 	&	0	&	3.74	&	0	&	1	&	5	&	7.92	&	0.119	\\
Education, Scientific Disciplines	&	 6,308 	&	 238 	&	0	&	3.72	&	0	&	1	&	5	&	8.72	&	0.110	\\
Computer Science, Information Systems	&	 45,865 	&	 715 	&	0	&	3.63	&	0	&	1	&	3	&	12.79	&	0.060	\\
Cultural Studies	&	 945 	&	 190 	&	0	&	3.55	&	0	&	1	&	4	&	9.13	&	0.297	\\
Engineering, Petroleum	&	 1,930 	&	 120 	&	0	&	3.53	&	0	&	1	&	4	&	6.63	&	0.151	\\
Robotics	&	 8,491 	&	 175 	&	0	&	3.38	&	0	&	1	&	4	&	7.58	&	0.082	\\
History \& Philosophy Of Science	&	 2,199 	&	 104 	&	0	&	3.35	&	0	&	2	&	4	&	5.96	&	0.127	\\
Computer Science, Hardware \& Architecture	&	 18,489 	&	 532 	&	0	&	3.34	&	0	&	1	&	3	&	10.96	&	0.081	\\
Mathematics	&	 25,450 	&	 253 	&	0	&	3.19	&	0	&	2	&	4	&	5.54	&	0.035	\\
Education \& Educational Research	&	 20,087 	&	 149 	&	0	&	3.14	&	0	&	1	&	4	&	6.85	&	0.048	\\
Area Studies	&	 2,046 	&	 44 	&	0	&	2.92	&	0	&	1	&	4	&	4.44	&	0.098	\\
Engineering, Mechanical	&	 50,972 	&	 262 	&	0	&	2.90	&	0	&	0	&	2	&	7.27	&	0.032	\\
Social Sciences, Interdisciplinary	&	 11,035 	&	 92 	&	0	&	2.73	&	0	&	0	&	3	&	5.62	&	0.053	\\
Film, Radio, Television	&	 398 	&	 31 	&	0	&	2.59	&	0	&	1	&	3	&	4.58	&	0.229	\\
History Of Social Sciences	&	 879 	&	 97 	&	0	&	2.49	&	0	&	1	&	3	&	4.24	&	0.143	\\
Computer Science, Theory \& Methods	&	 55,591 	&	 737 	&	0	&	2.45	&	0	&	1	&	2	&	8.79	&	0.037	\\
Art	&	 725 	&	 89 	&	0	&	2.33	&	0	&	1	&	3	&	5.55	&	0.206	\\
Logic	&	 1,786 	&	 38 	&	0	&	2.33	&	0	&	1	&	3	&	3.28	&	0.078	\\
Engineering, Marine	&	 2,110 	&	 101 	&	0	&	2.27	&	0	&	0	&	2	&	5.12	&	0.112	\\
Philosophy	&	 3,657 	&	 61 	&	0	&	2.22	&	0	&	1	&	3	&	4.06	&	0.067	\\
Language \& Linguistics	&	 5,174 	&	 112 	&	0	&	2.20	&	0	&	1	&	3	&	4.25	&	0.059	\\
Psychology, Psychoanalysis	&	 345 	&	 40 	&	0	&	2.17	&	0	&	1	&	3	&	4.04	&	0.218	\\
Music	&	 888 	&	 28 	&	0	&	2.06	&	0	&	1	&	3	&	3.10	&	0.104	\\
Religion	&	 2,335 	&	 81 	&	0	&	1.70	&	0	&	1	&	2	&	3.52	&	0.073	\\
History	&	 3,487 	&	 97 	&	0	&	1.53	&	0	&	1	&	2	&	2.84	&	0.048	\\
Asian Studies	&	 877 	&	 32 	&	0	&	1.10	&	0	&	0	&	1	&	1.99	&	0.067	\\
Literature, African, Australian, Canadian	&	 59 	&	 6 	&	0	&	1.07	&	0	&	1	&	1.5	&	1.26	&	0.164	\\
Humanities, Multidisciplinary	&	 2,559 	&	 53 	&	0	&	1.02	&	0	&	0	&	1	&	2.83	&	0.056	\\
Architecture	&	 1,376 	&	 145 	&	0	&	1.00	&	0	&	0	&	1	&	4.45	&	0.120	\\
Literature	&	 1,608 	&	 18 	&	0	&	0.81	&	0	&	0	&	1	&	1.77	&	0.044	\\
Theater	&	 300 	&	 9 	&	0	&	0.77	&	0	&	0	&	1	&	1.19	&	0.069	\\
Medieval \& Renaissance Studies	&	 485 	&	 11 	&	0	&	0.68	&	0	&	0	&	1	&	1.16	&	0.053	\\
Literature, American	&	 75 	&	 5 	&	0	&	0.68	&	0	&	0	&	1	&	1.02	&	0.117	\\
Classics	&	 325 	&	 9 	&	0	&	0.66	&	0	&	0	&	1	&	1.15	&	0.064	\\
Folklore	&	 134 	&	 7 	&	0	&	0.65	&	0	&	0	&	1	&	1.26	&	0.109	\\
Dance	&	 74 	&	 8 	&	0	&	0.55	&	0	&	0	&	0	&	1.30	&	0.152	\\
Literature, Romance	&	 269 	&	 6 	&	0	&	0.49	&	0	&	0	&	1	&	0.92	&	0.056	\\
Literature, British Isles	&	 220 	&	 4 	&	0	&	0.43	&	0	&	0	&	1	&	0.74	&	0.050	\\
Literary Theory \& Criticism	&	 498 	&	 45 	&	0	&	0.39	&	0	&	0	&	0	&	2.34	&	0.105	\\
Poetry	&	 42 	&	 3 	&	0	&	0.38	&	0	&	0	&	1	&	0.70	&	0.108	\\
Literature, German, Dutch, Scandinavian	&	 128 	&	 4 	&	0	&	0.36	&	0	&	0	&	1	&	0.70	&	0.061	\\
Literature, Slavic	&	 35 	&	 3 	&	0	&	0.29	&	0	&	0	&	0	&	0.67	&	0.113	\\
Literary Reviews	&	 35 	&	 2 	&	0	&	0.17	&	0	&	0	&	0	&	0.51	&	0.087	\\\hline

\end{longtable}

\end{document}